\documentclass[sigconf]{acmart}

\AtBeginDocument{%
  }

\usepackage{booktabs}
\usepackage{multirow}
\usepackage[normalem]{ulem}
\useunder{\uline}{\ul}{}

\usepackage{xcolor}
\newcommand\wu[1]{\textcolor{black}{#1}}

\renewcommand{\shortauthors}{}
\renewcommand\footnotetextcopyrightpermission[1]{}
\setcopyright{none}

\begin{document}

\title{MillGNN: Learning Multi-Scale Lead-Lag Dependencies for Multi-Variate Time Series Forecasting}

\author{Binqing Wu}
\affiliation{%
  \institution{College of Computer Science and Technology, \\ Zhejiang University}
  \city{Hangzhou}
  \state{Zhejiang}
  \country{China}
}
\email{binqingwu@cs.zju.edu.cn}

\author{Zongjiang Shang}
\affiliation{%
  \institution{College of Computer Science and Technology, \\ Zhejiang University}
  \city{Hangzhou}
  \state{Zhejiang}
  \country{China}
}
\email{zongjiangshang@cs.zju.edu.cn}

\author{Jianlong Huang}
\affiliation{%
  \institution{College of Computer Science and Technology, \\ Zhejiang University}
  \city{Hangzhou}
  \state{Zhejiang}
  \country{China}
}
\email{22251226@cs.zju.edu.cn}

\author{Ling Chen}
\authornote{Corresponding author.}
\affiliation{%
  \institution{College of Computer Science and Technology, \\ Zhejiang University}
  \city{Hangzhou}
  \state{Zhejiang}
  \country{China}
}
\email{lingchen@cs.zju.edu.cn}

\begin{abstract}
Multi-variate time series (MTS) forecasting is crucial for various applications. Existing methods have shown promising results owing to their strong ability to capture intra- and inter-variate dependencies. However, these methods often overlook lead-lag dependencies at multiple grouping scales, failing to capture hierarchical lead-lag effects in complex systems.
To this end, we propose MillGNN, a novel \underline{g}raph \underline{n}eural \underline{n}etwork-based method that learns \underline{m}ult\underline{i}ple grouping scale \underline{l}ead-\underline{l}ag dependencies for MTS forecasting, which can comprehensively capture lead-lag effects considering variate-wise and group-wise dynamics and decays.
Specifically, MillGNN introduces two key innovations: (1) a scale-specific lead-lag graph learning module that integrates cross-correlation coefficients and dynamic decaying features derived from real-time inputs and time lags to learn lead-lag dependencies for each scale, which can model evolving lead-lag dependencies with statistical interpretability and data-driven flexibility;
(2) a hierarchical lead-lag message passing module that passes lead-lag messages at multiple grouping scales in a structured way to simultaneously propagate intra- and inter-scale lead-lag effects, which can capture multi-scale lead-lag effects with a balance of comprehensiveness and efficiency. 
Experimental results on 11 datasets demonstrate the superiority of MillGNN for long-term and short-term MTS forecasting, compared with 16 state-of-the-art methods. 
\end{abstract}


\maketitle

\renewcommand{\shortauthors}{}
\settopmatter{printacmref=false}
\renewcommand\footnotetextcopyrightpermission[1]{}
\setcopyright{none}

\makeatletter
\let\ps@firstpagestyle\ps@empty
\let\ps@standardpagestyle\ps@empty
\makeatother
\pagestyle{empty}

\section{Introduction}

\begin{figure}
    \centering
    \includegraphics[width=0.8\linewidth]{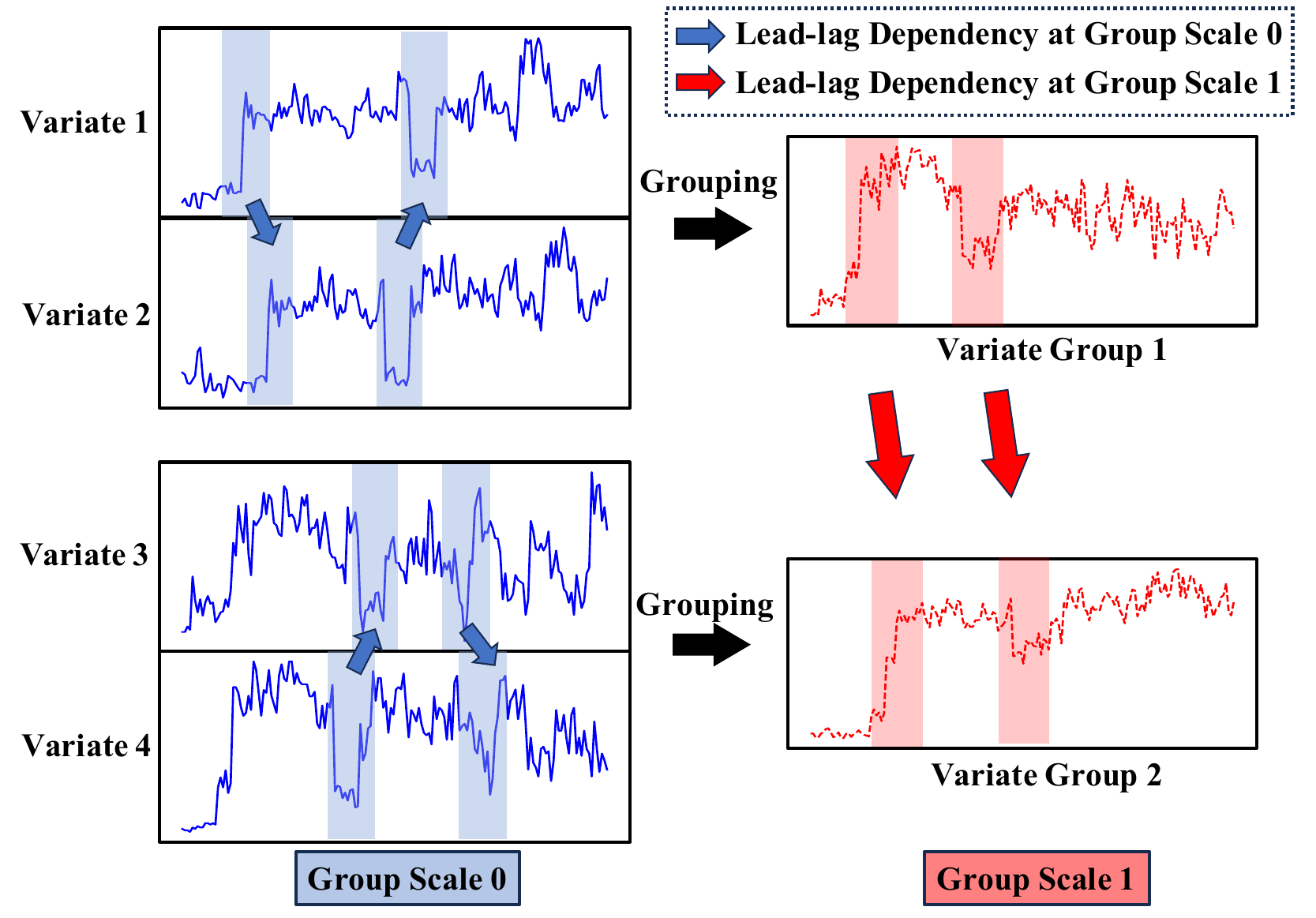}
    \caption{Illustration of lead-lag dependencies at different grouping scales.}
    \label{fig:intro}
\end{figure}

Multi-variate time series (MTS) forecasting, one of the most popular research topics, is a fundamental task in various domains, e.g., traffic management \cite{sun2024modwavemlp} and weather prediction \cite{wu2023weathergnn}. Due to the intricate intra-variate and inter-variate dependencies inherent in MTS, accurate MTS forecasting poses a significant challenge \cite{magnn}.

Recently, many deep learning methods have emerged for MTS forecasting, demonstrating remarkable performance. These methods can be mainly categorized into two groups: Variate-independent and variate-dependent. Variate-independent methods forecast each time series based on its historical values. They commonly employ Recurrent Neural Networks (RNNs) \cite{huang2024itrendrnn}, Convolutional Neural Networks (CNNs) \cite{wu2022timesnet}, attention mechanisms \cite{chen2024multi}, and Multi-layer Perceptrons (MLPs) \cite{wang2024timemixer} to capture intra-variate dependencies. Nevertheless, these methods disregard information from other variates. In contrast, variate-dependent methods forecast MTS by considering the historical values of each variate and those of other variates. They adopt attention mechanisms \cite{liuitransformer}, MLPs \cite{ekambaram2023tsmixer}, and Graph Neural Networks (GNNs) \cite{cai2024msgnet,wu2023dstcgcn} to incorporate features from multiple variates through implicit or explicit inter-variate dependencies. Nevertheless, these methods only capture inter-variate dependencies at the same time steps, often overlooking lead-lag dependencies, where segments of one time series (leading variate) affect another time series (lagging variate) with time lags \cite{brooks2001trading}.

Some deep learning methods have attempted to model lead-lag dependencies via pre-calculated cross-correlation coefficients \cite{zhao2024rethinking} and learnable methods, e.g., attentions \cite{pdformer} and GNNs \cite{long2024unveiling}. 
However, existing methods only capture lead-lag dependencies at the level of individual variates. They overlook a critical aspect: lead-lag dependencies often emerge at multiple grouping scales, i.e., between individual variates and between variate groups with different sizes, where each reflects distinct propagation dynamics in complex systems. For example, in power grid monitoring, a voltage drop at a specific transformer (a single sensor reading) might immediately affect the current readings at a directly connected substation, indicating a fine-grained lead-lag dependency between individual variates. A regional blackout in one part of the grid (e.g., a city district) may lead to load transfer and voltage instability across other distant regions over tens of minutes or hours, implying a coarse-grained lead-lag dependency between large variate groups. Consequently, capturing lead-lag dependencies at \textbf{multiple grouping scales (abbr. multi-scale)} is essential for accurately modeling system-wide hierarchical propagation behaviors that cannot be explained by single-scale interactions alone.

Despite their importance, learning and leveraging multi-scale lead-lag dependencies remains a non-trivial task due to two fundamental difficulties:
(1) \textbf{Scale variability}: Lead-lag dependencies are different at different levels of variate groups. As shown in Fig. \ref{fig:intro}, at coarser grouping scales, fine-grained fluctuations tend to be smoothed out, resulting in more stable and robust lead-lag patterns. In contrast, at finer scales, lead-lag dependencies become highly sensitive to local variations and are often dynamic, requiring models to adapt to fast-changing signals. This variability poses a challenge for unified modeling and calls for scale-specific dependency learning mechanisms.
% that can flexibly capture stable patterns at high levels and dynamic interactions at low levels.
(2) \textbf{Scale interaction}: Lead-lag effects, driven by lead-lag dependencies, propagate not only within but also across grouping scales. For example, traffic control in one area may affect neighboring areas and trigger cascading effects at smaller scales, e.g., causing road congestion in nearby areas. However, naive fusion across scales often results in redundant representations and incurs a high computational cost. This highlights the need for structured interaction mechanisms.

To this end, we propose MillGNN, a \underline{GNN}-based method to learn \underline{m}ult\underline{i}ple grouping scale \underline{l}ead-\underline{l}ag dependencies for MTS forecasting. To the best of our knowledge, MillGNN is the first work to learn and utilize multi-scale lead-lag dependencies for MTS forecasting, \wu{which can comprehensively capture lead-lag effects considering variate-wise and group-wise dynamics and decays.}
The main contributions are as follows:
\begin{itemize}
    \item For scale variability, we introduce a \underline{s}cale-specif\underline{i}c \underline{l}ead-\underline{l}ag \underline{g}raph \underline{l}earning (\underline{SiLL-GL}) module to learn lead-lag dependencies on each grouping scale. 
    This module first identifies initial lead-lag dependencies based on fast Fourier transform (FFT)-based cross-correlation calculation, and then models the evolution of these dependencies by integrating time-varying inputs and time lags via a decay-aware attention mechanism. The SiLL-GL module can model evolving lead-lag dependencies for each scale, adapting to their inherent scale-specific dynamics and decaying nature, while maintaining statistical interpretability and data-driven flexibility.
    \item For scale interaction, we introduce a \underline{hi}erarchical \underline{l}ead-\underline{l}ag \underline{m}essage \underline{p}assing (\underline{HiLL-MP}) module to propagate intra- and inter-scale lead-lag effects in a structure way.
    This module directly passes fined-grained lead-lag messages between similar variates/variate groups within each scale, and progressively passes coarsen-grained lead-lag messages between dissimilar variates/variate groups across scales. The HiLL-MP module can capture lead-lag effects with mixed granularities, achieving a balance between comprehensiveness and efficiency.
    \item We evaluate MillGNN on 11 real-world datasets, and the extensive experimental results demonstrate the superiority of MillGNN for long-term and short-term MTS forecasting, compared with 16 state-of-the-art (SOTA) methods. 
\end{itemize}

\section{Related Works}
\subsection{Multi-Variate Time Series Forecasting}
Many methods have emerged for MTS forecasting, ranging from statistics methods \cite{arima,svr} to deep learning methods \cite{mtgnn,zhou2021informer,AdaMSHyper,shang2024mshyper}. Due to the strong representation ability, deep learning methods have become mainstream and can be mainly categorized into two groups: Variate-independent and variate-dependent. 
Variate-independent methods forecast each time series based on its historical values. They commonly utilize RNNs \cite{salinas2020deepar,lin2023segrnn}, CNNs \cite{wu2022timesnet,li2024lagcnn}, attention mechanisms  \cite{zhou2022fedformer,nie2022time,piao2024fredformer}, and MLPs \cite{zeng2023transformers,wang2024timemixer,lincyclenet} to capture intra-variate dependencies. For example, PatchTST \cite{nie2022time} utilizes patching and channel-independent attentions to model point-level and subseries-level intra-variate dependencies, respectively. TimeMixer \cite{wang2024timemixer} adopts seasonal-trend decomposition and utilizes MLPs to model multi-scale intra-variate dependencies. 
Despite their successes, these methods disregard information from other variates, encountering information limitations. 
Variate-dependent methods consider the historical values of multiple variates. They integrate features between variates using attention mechanisms  \cite{zhang2023Crossformer,liuitransformer,wu2025revisiting}, MLPs \cite{ekambaram2023tsmixer,huang2024hdmixer,tian2025hypermixer}, GNNs \cite{mtgnn,huang2023crossgnn,cai2024msgnet,wu2025sthyper}, and FFT-related technologies \cite{wang2025filterts}.
For example, iTransformer \cite{liuitransformer} adopts the attention on the variate dimension to capture inter-variate dependencies. MSGNet \cite{cai2024msgnet} constructs graphs based on scale-aware embeddings to capture inter-variate dependencies at different time scales. 
\wu{FilterTS\cite{wang2025filterts} detects dataset‑wide stable frequencies and uses the other variates as adaptive filters to amplify these shared components across scales.}
Nevertheless, these methods overlook lead-lag dependencies, which may fail to capture lead-lag effects among variates.

\subsection{Lead-Lag Dependency Learning for Time Series}
Lead-lag dependencies refer to dependencies where the segments of one time series (leading variate) affect another time series (lagging variate) after certain time lags \cite{brooks2001trading}.
Such dependencies have been investigated in various fields, e.g., economy \cite{yichi2023dynamic}, energy \cite{ZHU2024123194}, and traffic \cite{pdformer}.
For MTS, some methods have attempted to model these dependencies implicitly. 
For example, PDFormer \cite{pdformer} adopts a learnable pattern set to memorize historical features and integrates historical and current features via attention for traffic flow forecasting. Since the pattern set is variate-shared, it can consider delays but cannot explicitly capture lead-lag dependencies between pairwise variates. 
Some methods have tried to model lead-lag dependencies explicitly.
For example, LIFT \cite{zhao2024rethinking}, playing as a plugin toolkit, pre-calculates cross-correlation between variates to capture locally stationary lead-lag dependencies.
STDDE \cite{long2024unveiling} utilizes cross-correlations to identify lead-lag dependencies and utilizes graph-enhanced neural differential equations to consider continuous lead-lag effects for traffic flow forecasting. 
VCformer \cite{yang2024vcformer} computes cross-correlations with different lags and aggregates different lags with learnable parameters to capture the lead-lag dependencies between variates. However, these methods only capture lead-lag dependencies between individual variates, overlooking those at multiple grouping scales. 

Therefore, we propose MillGNN, the first work to learn and utilize multi-scale lead-lag dependencies for MTS forecasting. In brief, MillGNN learns evolving lead-lag dependencies based on statistical and learnable features and propagates intra- and inter-scale lead-lag effects in a structured way, which can comprehensively deal with lead-lag effects between individual variates and between variate groups.

\begin{figure*}
    \centering
    \includegraphics[width= 0.7 \linewidth]{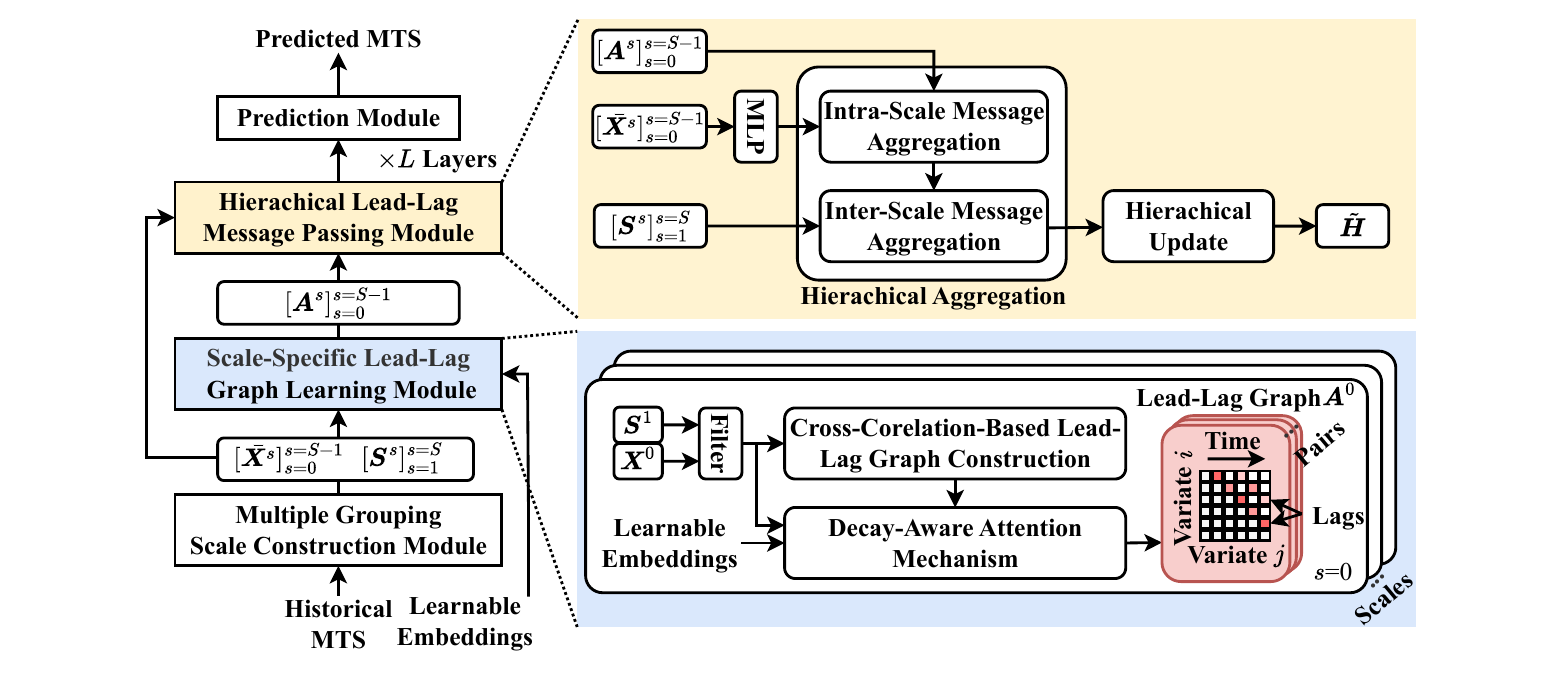}
    \caption{Framework of MillGNN.}
    \label{fig:framework}
\end{figure*}

\section{Preliminaries}
\subsection{Task Formulation}
In MTS, each time series represents a variate. Given historical MTS $\boldsymbol{X}=\{\boldsymbol{X}_i\}_{i=1}^N \in \mathbb{R}^{N \times L}$, where $N$ is the number of variates and $L$ is the input length, the goal of MTS forecasting is to learn a function that maps the historical MTS with $L$ time steps to the future MTS with $H$ time steps. This process is formulated as:
\begin{equation}
    \hat{\boldsymbol{X}}=f(\boldsymbol{X} ; \boldsymbol{\Theta}),
\end{equation}
where $\boldsymbol{\Theta}$ denotes all learnable parameters of $f$, and $\hat{\boldsymbol{X}} \in \mathbb{R}^{N \times H}$ is the predicted results.

\subsection{Lead-Lag Graph Definition}\label{llgdef}
A lead-lag graph for MTS is defined as $\mathcal{G} = (\boldsymbol{V}, \boldsymbol{E}, \boldsymbol{A})$, where $\boldsymbol{V}$ represents the set of nodes, $\boldsymbol{E}$ denotes the set of directed edges, and $\boldsymbol{A}$ is the weighted adjacency matrix. Given MTS $\boldsymbol{X} \in \mathbb{R}^{N \times L}$, where ${N}$ is the number of variates and ${L}$ is the input length, the node set is defined as $\boldsymbol{V}=\left\{v_{i, n} \mid i \in[N], n \in[L]\right\}$, where $v_{i,n}$ represents the $n$-th time step of variate $i$, resulting in $|\boldsymbol{V}| = N \times L$. The edge set is defined as $\boldsymbol{E} = \left\{\varepsilon_{ij,nm} \mid i,j \in[N], n,m \in[L]\right\}$, where $\varepsilon_{ij,nm}$ represents the lead-lag dependency from $v_{i,n}$ to $v_{j,m}$. The time lag associated with this dependency is $m-n$ ($m \geq n$). The weighted adjacency matrix $\boldsymbol{A} \in \mathbb{R}^{N \times N \times L \times L}$ quantifies the strength of lead-lag dependencies, with $\boldsymbol{A}_{ij, nm} \neq 0 \Longrightarrow \varepsilon_{ij,nm} \in \boldsymbol{E}$. Crucially, due to the causality constraint, i.e., only the past time steps can affect current or future time steps, the submatrix $\boldsymbol{A}_{ij} \in \mathbb{R}^{L \times L}$ for each variate pair must satisfy the condition $\boldsymbol{A}_{ij, nm}=0 \quad$ for $m<n$. \textbf{This implies that $\boldsymbol{A}_{ij}$ is upper triangular for all variate pairs.} For simplicity, we can utilize $\boldsymbol{A}$ to represent a lead-lag graph.

\subsection{FFT-Based Cross-Correlation Calculation} 
The cross-correlation coefficient is essential to detect lead-lag dependencies. To improve computational efficiency, an extension of the Wiener–Khinchin theorem \cite{wiener1930generalized} has been employed to estimate cross-correlation coefficients \cite{zhao2024rethinking,yang2024vcformer}.
In brief, by leveraging FFT, the convolution performed in the time domain can be transformed into element-wise multiplication in the frequency domain. This transformation reduces the computational complexity of calculating the cross-correlation between two time series from $O(L^2)$ to $O(L \log L)$, where $L$ is the length of the time series. The FFT-based cross-correlation calculation is formulated as:
\begin{equation} \label{cross}
\{R_{ij}(\tau)\}_{\tau=0}^{L-1} = \operatorname{Corr}(X_{i}, X_{j}) = \frac{1}{L} \mathcal{F}^{-1}(\mathcal{F}(X_{j}) \odot \overline{\mathcal{F}(X_{i})}),
\end{equation}
where $\mathcal{F}$ is the FFT, $\mathcal{F}^{-1}$ is the inverse FFT, $\odot$ is the element-wise product, and $\overline{(\cdot)}$ is the conjugate operation. $\{R_{ij}(\tau)\}_{\tau=0}^{L-1}$ is the set of the estimated coefficients for all possible time lags $\tau$ between time series $i$ and $j$.

\section{Methodology}

The framework of MillGNN is illustrated in Fig. \ref{fig:framework}.
MillGNN first groups variates based on their temporal similarities to construct multiple grouping scales. Next, the SiLL-GL module is introduced to model scale-specific lead-lag dependencies. On each scale, the FFT-based cross-correlation calculation is conducted to construct an initial lead-lag graph. This graph serves as the foundation for further refinement. MillGNN then applies a decay-aware attention mechanism to dynamically adjust the edge weights of the initial lead-lag graph, adapting to group-wise dynamics and decays.
Given the learned multi-scale lead-lag graphs, the HiLL-MP module is introduced to propagate intra- and inter-scale lead-lag effects. It directly passes fined-grained lead-lag messages between similar variates/variate groups within each scale, and progressively passes coarsen-grained lead-lag messages between dissimilar variates/variate groups across scales.
Finally, MillGNN employs linear projections to predict MTS.

\subsection{Multiple Grouping Scale Construction} \label{hierachy}
Since lead-lag dependencies often occur between variates with high similarity \cite{zhao2024rethinking}, we first measure the similarity between variates using Dynamic Time Warping, due to its robustness to temporal misalignment, as supported by lead-lag related prior work \cite{mu2025gaussian,li2025follow}.
This computation is performed using all MTS in the training set. We binarize the similarity values, obtaining a similarity graph $\boldsymbol{D}^0 \in \{0,1\}^{N \times N}$, where $N$ represents the number of variates. The element $\boldsymbol{D}^0_{ij} = 1$ indicates a significant similarity between variates $i$ and $j$. We group variates based on $\boldsymbol{D}^0$, ensuring that intra-group similarity is stronger than inter-group similarity. For each grouping scale $s$ ($s > 0$), this process is formulated as:
\begin{equation}
    \boldsymbol{S}^s = \operatorname{Clustering}(N^s, \boldsymbol{D}^{s-1}),
\end{equation}
where $N^s$ denotes the number of variate groups at scale $s$. The clustering algorithm is flexible and can be changed based on specific requirements. We explore multiple choices in Section 5.3. The assignment matrix $\boldsymbol{S}^s \in \{0,1\}^{N^{s-1} \times N^s}$ indicates the assignment of groups from scale $s-1$ to scale $s$. At coarser scales, each group encompasses a larger receptive field and can aggregate information from a larger number of variates. After that, the similarity graph and corresponding MTS at scale $s$ ($s > 0$) are formulated as: 
\begin{equation}
    \boldsymbol{D}^s = (\boldsymbol{S}^s)^T \boldsymbol{D}^{s-1} \boldsymbol{S}^s,
    \boldsymbol{X}^s = (\boldsymbol{S}^s)^T \boldsymbol{X}^{s-1}
\end{equation}
where $\boldsymbol{D}^s \in \mathbb{R}^{N^s \times N^s}$ indicates the similarities between groups. $\boldsymbol{X}^s \in \mathbb{R}^{N^s \times L}$ are the corresponding MTS. 

Since lead-lag dependencies often occur between segments, $\boldsymbol{X}^s \in \mathbb{R}^{N^s \times L}$ is divided into patches. Larger patch lengths are applied at coarser scales to capture lead-lag dependencies over a broader space-time scope. After patchifying, the multi-scale MTS of $S$ grouping scales is transformed into $\{\bar{\boldsymbol{X}}^s\}_{s=0}^{S-1} = \{\bar{\boldsymbol{X}}^0, \bar{\boldsymbol{X}}^1, \cdots, \bar{\boldsymbol{X}}^{S-1} \}$, and $\bar{\boldsymbol{X}}^s \in \mathbb{R}^{N^s \times P^s \times p^s}$, where $P^s = L/p^s$ and $p^s$ are the patch number and patch length of scale $s$, respectively. 

At this stage, we construct a hierarchy of time series $\{\bar{\boldsymbol{X}}^s\}_{s=0}^{S-1}$ and their corresponding similarity graphs $\{\boldsymbol{D}^s\}_{s=0}^{S-1}$ on multiple grouping scales. When $s=0$, each variate forms an individual group, representing the original variate level. For $s > 0$, the representations reflect coarser groupings of variates at increasing scales. The assignment matrices $\{\boldsymbol{S}^s\}_{s=1}^{S-1}$ specify the grouping structure at each scale, mapping finer groups to coarser ones.

\subsection{Scale-Specific Lead-Lag Graph Learning}
To address scale variability, the SiLL-GL module is introduced to learn lead-lag dependencies for each scale. It involves the cross-correlation-based lead-lag graph construction and a decay-aware attention mechanism, enabling interpretable and flexible modeling of scale-specific asynchronous temporal dynamics.

\subsubsection{Cross-Correlation-Based Lead-Lag Graph Construction}
For each scale $s$, lead-lag dependencies are captured between pairs of groups that share the same parent group at scale $s+1$, as these pairs exhibit high similarity. Thus, we filter group pairs satisfying this condition as candidates.
For two groups $i$ and $j$ at scale $s$ in candidates, average pooling is first applied within each patch to reduce non-stationary impact, e.g., noise. Then, we conduct the FFT-based cross-correlation calculation (Eq. \ref{cross}) and select $K$ time lags with the highest Top-$K$ cross-correlation coefficients. Given the set of selected time lags $\boldsymbol{\tau}_{ij}$, where $|\boldsymbol{\tau}_{ij}| = K$, the initial lead-lag graph $\boldsymbol{C}_{ij}^s \in \mathbb{R}^{P^s \times P^s}$ between groups $i$ and $j$ at scale $s$ is formulated as:
\begin{equation}
\boldsymbol{C}_{ij,nm}^s= \begin{cases}1, & m-n \in \boldsymbol{\tau}_{ij}  \text { and } n < P^s-\max \left(\boldsymbol{\tau}_{ij} \right) \\ 0, & \text {otherwise}\end{cases}
\end{equation}
where $m-n = \tau^k_{ij} \in \boldsymbol{\tau}_{ij}$ means there is an initial lead-lag dependency from the $n$-th time step of $i$ to the $m$-th time step of $j$ with the time lag (offset) $\tau^k_{ij}$. $n < P^s-\max \left(\boldsymbol{\tau}\right)$ ensures that the matrix is within the valid range. 
According to Section \ref{llgdef}, $\boldsymbol{C}_{ij}^s$ is upper triangular for all $i,j \in [N^s]$.

At scale $s$, the constructed initial lead-lag graph for all groups $\boldsymbol{C}^s$ is in the shape of $\{0,1\}^{N^s \times N^s \times P^s \times P^s}$, where each node is a patch of a group, and each edge is a lead-lag dependency between patches.

\subsubsection{\wu{Decay-Aware Attention Mechanism}}\label{decay}
To capture the dynamic and decaying nature of lead-lag dependencies, we design a decay-aware attention mechanism to explicitly model the diminishing influence over time.
\wu{Modeling decay is non-trivial due to the coexistence of two competing effects, i.e., rapid dissipation and periodic revival.
For example, when groups are weakly related, their mutual influence tends to decay rapidly, intensifying the decay. In contrast, cyclical or seasonal processes can intermittently revive past patterns, offsetting the decay. Therefore, our mechanism introduces the excitatory and inhibitory rates to model such two competing effects, which can capture decay effects more comprehensively.} 

For scale $s$, we first introduce two group embedding $\boldsymbol{E}_{\text{Ni}}^s \in \mathbb{R}^{N^s \times d_{\text{e}}}$, $\boldsymbol{E}_{\text{Ne}}^s \in \mathbb{R}^{N^s \times d_{\text{e}}}$ to capture the excitatory and inhibitory characteristics of groups. In addition, we introduce one time embedding $\boldsymbol{E}_{\text{P}}^s \in \mathbb{R}^{P^s \times d_{\text{e}}}$ to represent the temporal characteristics at each patch position. We combine these embeddings as:
\begin{equation}
\boldsymbol{E}_{\text{i}}^{s} = \boldsymbol{E}_{\text{Ni}}^{s} \oplus\boldsymbol{E}_{\text{P}}^{s},
\boldsymbol{E}_{\text{e}}^{s} = \boldsymbol{E}_{\text{Ne}}^{s} \oplus\boldsymbol{E}_{\text{P}}^{s}
\end{equation}
where $\oplus$ denotes the addition with broadcasting. 
$\boldsymbol{E}_{\text{i}}^{s},\boldsymbol{E}_{\text{e}}^{s} \in \mathbb{R}^{N^s \times P^s \times d_{\text{e}}}$ are two embeddings of inherent excitatory and inhibitory characteristics at each patch position. Consequently, the excitatory and inhibitory rates between patches can be formulated as:
\begin{equation}
    \boldsymbol{\beta}_{\text{e}}^{s} = \operatorname{SoftMax}(\boldsymbol{E}_{\text{e}}^{s}(\boldsymbol{E}_{\text{e}}^{s})^T),
    \boldsymbol{\beta}_{\text{i}}^{s} = \operatorname{SoftMax}(\boldsymbol{E}_{\text{i}}^{s}(\boldsymbol{E}_{\text{i}}^{s})^T)
\end{equation}
where $\boldsymbol{\beta}_{\text{e}}^{s} \in \mathbb{R}^{N^s \times N^s \times P^s \times P^s}$ and $\boldsymbol{\beta}_{\text{i}}^{s} \in \mathbb{R}^{N^s \times N^s \times P^s \times P^s}$ are optimized during training and shared across all input samples at inference.

To account for temporal dynamics, we further extract real-time input features via attention, formulated as:
\begin{equation}
\begin{aligned}
     \boldsymbol{Q}^s = \operatorname{Linear}(\bar{\boldsymbol{X}}^s), \boldsymbol{K}^s = \operatorname{Linear}(\bar{\boldsymbol{X}}^s),\boldsymbol{\alpha}^{s} = \operatorname{SoftMax}(\boldsymbol{Q}^s (\boldsymbol{K}^s)^T/\sqrt{d_{\text{a}}})
\end{aligned}
\end{equation}
where $d_{\text{a}}$ is the hidden dimension of the query $\boldsymbol{Q}^s \in \mathbb{R}^{N^s \times P^s \times d_{\text{a}}}$ and the key $\boldsymbol{K}^s \in \mathbb{R}^{N^s \times P^s \times d_{\text{a}}}$. Both the query and key are derived from the shared $\bar{\boldsymbol{X}}^s$ through linear projections.

Then, we integrate the excitatory and inhibitory rates with the attention scores to formulate the dynamic weights considering decaying effects:
\begin{equation}
     \boldsymbol{\Lambda}^{s} = \operatorname{ReLU}(
     \underbrace{e^{-\boldsymbol{\beta}_{\text{e}}^{s} \boldsymbol{\Delta}^s}}_{\text{Excitatory}}
     + \underbrace{(-(1-\boldsymbol{\alpha}^{s}) e^{-\boldsymbol{\beta}_{\text{i}}^{s}\boldsymbol{\Delta}^s})}
     _{\text{Inhibitory}})
\end{equation}
where $e$ denotes the exponential kernel, and $\boldsymbol{\Delta}^s \in \mathbb{R}^{N^s \times N^s \times P^s \times P^s}$ is the time‑lag tensor that stores the intervals between patch pairs.
Larger excitatory rates $\boldsymbol{\beta}_{\text{e}}(0<\boldsymbol{\beta}_{\text{e}}<1)$ shrink the value of the excitatory term, producing a stronger decay. Conversely, larger dynamic attention scores $\boldsymbol{\alpha}^{s}(0<\boldsymbol{\alpha}^{s}<1)$ and larger inhibitory rates $\boldsymbol{\beta}_{\text{i}}(0<\boldsymbol{\beta}_{\text{i}}<1)$ enlarge the values of inhibitory term, yielding a weaker decay.

To this end, for scale $s$, the lead-lag graph with dynamic and decaying weights can be formulated as:
\begin{equation}
\boldsymbol{A}^{s} = \boldsymbol{C}^s \odot \boldsymbol{\Lambda}^s
\end{equation}
where $\boldsymbol{A}^{s}$ is in the shape of $\mathbb{R}^{N^s \times N^s \times P^s \times P^s}$, encoding the lead–lag strength for every patch pair at scale $s$.

% We adopt the logits warper softmax \cite{huang2023crossgnn} on $\boldsymbol{A}^{s}$ that only conducts the softmax function with the dependencies existing in the lead-lag graph. 

For all scales, we can obtain the lead-lag graphs $\{\boldsymbol{A}^s\}_{s=0}^{S-1}$, which represent the lead-lag dependencies fully considering the variate-wise and group-wise dynamics and decays.

% time-aligned and within the same group.

\subsection{Hierarchical Lead-Lag Message Passing}
To address scale interactions, the HiLL-MP module is introduced to propagate intra-and inter-scale lead-lag effects. It involves hierarchical aggregation and hierarchical update operations.

\subsubsection{Hierarchical Aggregation Operation}

\begin{figure}
    \centering
    \includegraphics[width= \linewidth]{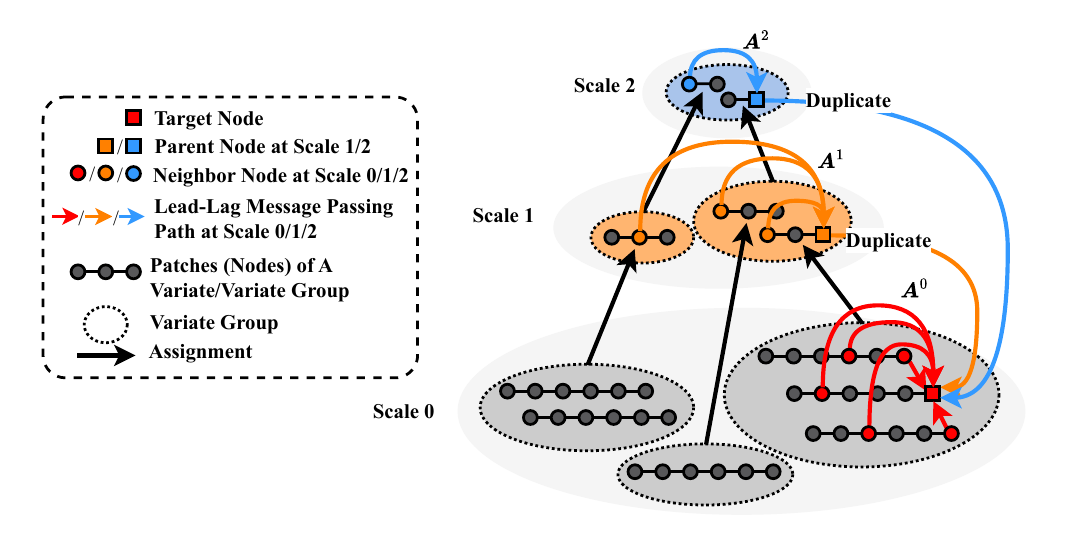}
    \caption{Illustration of hierarchical message aggregation.}
    \label{fig:method_case}
\end{figure}

The main idea of the hierarchical aggregation is to coarsen and aggregate messages based on similarity.
By leveraging the similarity hierarchy discussed in Section \ref{hierachy}, this operation aggregates fined-grained lead-lag messages between similar groups and coarsen-grained lead-lag messages between dissimilar groups.

Since MTS forecasting focuses on variate-wise prediction, we aggregate messages from groups to individual variates at scale $s=0$. Given the assignment matrices $\{\boldsymbol{S}^s\}_{s=1}^S$ and multi-scale lead-lag graphs $\{\boldsymbol{A}^s\}_{s=0}^{S-1}$, the hierarchical aggregation is formulated as:
\begin{equation} \label{message}
\begin{aligned}
     \boldsymbol{M} & = \underbrace{\sigma(\boldsymbol{A}^0 \boldsymbol{H}^0 \boldsymbol{\theta}^0)}_{\text{Intra-scale}}
     + \underbrace{\sum_{s=1}^{S-1} \sigma(\frac{1}{|v_s|} \prod_{l=1}^{s} \boldsymbol{S}^l \bigotimes (\boldsymbol{A}^s \boldsymbol{H}^{s}\boldsymbol{\theta}^s))}_{\text{Inter-scale}} 
\end{aligned}
\end{equation}
where $\sigma$ is the activation function, and $\boldsymbol{H}^{s} = \operatorname{MLP}(\bar{\boldsymbol{X}}^s) \in \mathbb{R}^{N^s \times P^s \times d}$. $\boldsymbol{A}^s \boldsymbol{H}^{s}$ represent the aggregated lead-lag messages at scale $s$ (multiplying with reshape). $\prod_{l=1}^{s} \boldsymbol{S}^l \in \{0,1\}^{N \times N^s}$ indicates the direct assignment between the groups at scale $s>0$ and variates at $s=0$. 
$|v_s|$ is the number of variates contained by groups at $s$, decided by each column of $\prod_{l=1}^{s} \boldsymbol{S}^l$. $\boldsymbol{\theta}^s$ are the corresponding parameters. $\bigotimes$ represents the duplication operation, which is essential to improve the efficiency of inter-scale message passing. Since $\prod_{l=1}^{s} \boldsymbol{S}^l$ is binary, messages from groups at scale $s$ to their containing variates are identical and can be \textbf{directly duplicated} to the containing variates after average instead of exhaustive multiplication. 

Fig. \ref{fig:method_case} provides an illustration for better understanding. A target node (a patch within a variate) receives two types of messages: intra-scale and inter-scale messages. Intra-scale messages are passed directly via lead-lag dependencies within the same group, as defined by $\boldsymbol{A}^0$. Inter-scale messages are first aggregated at the parent nodes (the nodes in the groups at higher scales that contain the target node) according to $\boldsymbol{A}^1$ and $\boldsymbol{A}^2$ and subsequently duplicated and assigned to the target node.

By integrating the similarity hierarchy and duplication operation, the exhaustive pairwise message passing between nodes within and across multiple scales is significantly reduced by coarsening and duplicating, which can aggregate intra- and inter-scale messages efficiently.

\subsubsection{Hierarchical Update Operation}
This operation first updates the features of nodes in variates and then updates the features of nodes in groups. The process is formulated as:
\begin{equation}
\begin{aligned}
    \text{Intra-scale: } &  \boldsymbol{H}^{\prime} = \sigma((\boldsymbol{H} ||\boldsymbol{M}) \cdot \boldsymbol{\theta}_{\text{update}}), \\
    \text{Inter-scale: } &  \boldsymbol{H}^{s \prime} = \sigma((\boldsymbol{H}^s || \operatorname{Pool}(\boldsymbol{H}^{\prime} \prod_{l=1}^{s} \boldsymbol{S}^l)) \cdot \boldsymbol{\theta}_{\text{update}}^s)
\end{aligned}
\end{equation}
where $||$ represents concatenation. $\boldsymbol{H}^{(\cdot)}$ and $\boldsymbol{H}^{(\cdot) \prime}$ are the original and updated features, respectively. $\boldsymbol{\theta}_{\text{update}}^{(\cdot)}$ are the corresponding parameters.

After one layer of the HiLL-MP module, we can propagate intra- and inter-scale lead-lag effects efficiently. This module can be stacked multiple layers with skip connections to further enhance propagation. We use the features of nodes in variates from the last layer, denoted as $\Tilde{\boldsymbol{H}} \in \mathbb{R}^{N \times P^0 \times d}$, as the final output.

\noindent \textbf{Proposition.} \textit{The complexity of the HiLL-MP module is $O(SLN)$, where $N$ is the number of variates, $L$ is the input length, and $S$ is the number of grouping scales.} 

\noindent \textbf{Proof.}
Since the multi-scale structure is constructed using grouping and pacifying operations, the number of nodes at any scale is strictly less than $N \times L$. Therefore, the total number of nodes $V$ is bound by \underline{$S \times N \times L$}.

% \textbf{Part 2.} \textit{The total number of edges in multi-scale lead-lag graphs, denoted as $|E_L|$, is linear with respect to $NL+N^2$.}

When $s=0$, the number of groups is $N$. When $s>0$, the number of groups at scale $s$ is equal to $N^{s} = \frac{N^{s-1}}{k} =  \frac{N}{k^{s}}$, where $k$ is the expected number of variables in each group, and $S = \mathrm{log}_k N +1$ when the number of groups at the largest scale is equal to 1. 
Within a single group, the group contains the expected $k$ variables/sub-groups, indicating $k^2$ pairs. For each pair, there are $\tau$ potential time lags, indicating at most $\tau$ edges for each time step and consequently $\tau \times L$ for $L$ time steps.
Therefore, for all groups at scale $s$, the edge number $E^s$ is $\tau L \times k^2 \times N/k^{s}$. For all scales, the total number of edges $E$ is formulated as:
$$
E = \sum_{s=0}^{S-1} E^s = \sum_{s=0}^{S-1}  \tau  L k^2 \times N/k^{s}=  \tau L k^2 N \sum_{s=0}^{S-1} \frac{1}{k^{s}}\ = \tau  L k^2 N \frac{1-(1 / k)^{S}}{1-1 / k}
$$

By substituting $S = \mathrm{log}_k N +1$, $E = \tau^{\prime} L \cdot \frac{k^3}{k-1} \cdot\left(N-\frac{1}{k}\right)$. Since $\tau^{\prime}$ and $k$ are constants, the number of edges $E$ is linear with respect to \underline{$N \times L$}. 

In general, the complexity of message passing with a graph is $O(V+E)$ \cite{hamilton2020graph}.
Therefore, the complexity of message passing \textit{within} scales is $O(SNL+ NL) =$ \underline{$O(SNL)$}.
For the complexity of message passing \textit{across} scales, since $\prod_{l=1}^{s} \boldsymbol{S}^l$ is binary, messages from groups at scale $s$ to their containing variates are identical and can be directly duplicated to the containing variates after average instead of exhaustive multiplication. The complexity is \underline{$O(1)$}.

To this end, the total complexity of the hierarchical lead-lag message passing mechanism is proved to be \underline{$O(SNL)$}, which is significantly lower than the complexity of adopting message passing across all nodes at all scales, i.e., $O((SNL)^2)$.

\subsection{Prediction}
Following PatchTST \cite{nie2022time}, we flatten the patch and hidden dimensions of $\Tilde{\boldsymbol{H}}$ and then project them to the output length using linear projections. The process is formulated as:
\begin{equation}
\begin{aligned}
    \hat{\boldsymbol{X}} = \operatorname{Linear}(\operatorname{Flatten}(\Tilde{\boldsymbol{H}}))
\end{aligned}
\end{equation}
where $\hat{\boldsymbol{X}} \in \mathbb{R}^{N \times H}$ is the predicted results. 
For the loss function, we choose Mean Squared Error (MSE) following existing works \cite{liuitransformer,wang2024timemixer}.

\section{Experiment}

\subsection{Experimental Settings}

\begin{table}[]
    \centering
    \caption{Dataset statistics.}
    \label{tab:dataset}
    \resizebox{\linewidth}{!}{
    \begin{tabular}{lcccccl}
    \hline
    {Datasets} & {Dataset Size} & {\# variates} & {Sample rate} & {Input length} & {Output length}  \\ 
    \hline
    ETTh1, ETTh2 & 8545, 2881, 2881 & 7 & 1 hour & 96 & 96, 192, 336, 720 \\
    ETTm1, ETTm2 & 34465, 11521, 11521 & 7 & 15 min & 96 & 96, 192, 336, 720 \\
    Weather & 36792, 5271, 10540 & 21 & 10 min &96 &96, 192, 336, 720  \\
    Electricity & 18317, 2633, 5261 & 321 &  1 hour &96  &96, 192, 336, 720  \\
    Exchange & 5120, 665, 1422 & 8 &  1 day &96  &96, 192, 336, 720  \\
    Traffic & 12185, 1757, 3509 & 862 &  1 hour &96  &96, 192, 336, 720  \\
    \hline
    PEMS04 & 10088, 3291, 3292 & 307 & 5 min & 96 & 12 \\
    PEMS08 & 10606, 3464, 3465 & 170 & 5 min & 96 & 12 \\
    China-AQI & 14161, 1920, 3962 & 209 & 1 hour & 96 & 24 \\
    \hline
    \end{tabular}}
\end{table}

\noindent \textbf{Datasets.} We choose \textbf{11 real-world datasets} for long-term and short-term MTS forecasting. The dataset statistics are shown in Table \ref{tab:dataset}. For long-term, we follow the configuration from iTransformer \cite{liuitransformer} and use ETT (ETTh1, ETTh2, ETTm1, ETTm2), Weather, Electricity, Exchange, and Traffic datasets. For short-term, we follow the configuration from TimeMixer \cite{wang2024timemixer} and use PEMS04 and PEMS08 datasets. In addition, we follow the configuration from GAGNN \cite{chen2023group} and use China-AQI dataset.

\begin{table*}[]
\centering
\caption{Long-term forecasting results with the input length $L=96$ and predicted lengths $H \in \{96,192,336,720\}$.}
\label{tab:long-term}
\resizebox{\textwidth}{!}{
\begin{tabular}{@{}cc|cc|cc|cc|cc|cc|cc|cc|cc|cc|cc|cc|cc|cc|cc|cc|cc|cc@{}}
\toprule
\multicolumn{2}{c|}{} & \multicolumn{2}{c}{\textbf{MillGNN}} & \multicolumn{2}{c}{\textbf{FilterTS}} & \multicolumn{2}{c}{\textbf{TimeMixer}} & \multicolumn{2}{c}{\textbf{LIFT}$^*$} & \multicolumn{2}{c}{\textbf{VCFormer}} & \multicolumn{2}{c}{\textbf{iTransformer}} & \multicolumn{2}{c}{\textbf{MSGNet}} & \multicolumn{2}{c}{\textbf{CrossGNN}} & \multicolumn{2}{c}{\textbf{PatchTST}} & \multicolumn{2}{c}{\textbf{Crossformer}} & \multicolumn{2}{c}{\textbf{TimesNet}} & \multicolumn{2}{c}{\textbf{DLinear}} & \multicolumn{2}{c}{\textbf{SCINet}} & \multicolumn{2}{c}{\textbf{FEDformer}} & \multicolumn{2}{c}{\textbf{Stationary}} & \multicolumn{2}{c}{\textbf{Autoformer}} & \multicolumn{2}{c}{\textbf{MTGNN}} \\

\multicolumn{2}{c|}{\multirow{-2}{*}{\textbf{Models}}} & \multicolumn{2}{c}{\textbf{(Ours)}} & \multicolumn{2}{c}{\textbf{2025}} &   \multicolumn{2}{c}{\textbf{2024}} &  \multicolumn{2}{c}{\textbf{2024}} & \multicolumn{2}{c}{\textbf{2024}} & \multicolumn{2}{c}{\textbf{2024}} & \multicolumn{2}{c}{\textbf{2024}} & \multicolumn{2}{c}{\textbf{2023}} & \multicolumn{2}{c}{\textbf{2023}} & \multicolumn{2}{c}{\textbf{2023}} & \multicolumn{2}{c}{\textbf{2023}} & \multicolumn{2}{c}{\textbf{2023}} & \multicolumn{2}{c}{\textbf{2022}} & \multicolumn{2}{c}{\textbf{2022}} & \multicolumn{2}{c}{\textbf{2022}} & \multicolumn{2}{c}{\textbf{2021}} & \multicolumn{2}{c}{\textbf{2020}} \\ \midrule

\multicolumn{2}{c|}{\textbf{Metric}} & \textbf{MSE} & \multicolumn{1}{c|}{\textbf{MAE}} & \textbf{MSE} & \multicolumn{1}{c|}{\textbf{MAE}} & \textbf{MSE} & \multicolumn{1}{c|}{\textbf{MAE}} & \textbf{MSE} & \multicolumn{1}{c|}{\textbf{MAE}} & \textbf{MSE} & \multicolumn{1}{c|}{\textbf{MAE}} & \textbf{MSE} & \multicolumn{1}{c|}{\textbf{MAE}} & \textbf{MSE} & \multicolumn{1}{c|}{\textbf{MAE}} & \textbf{MSE} & \multicolumn{1}{c|}{\textbf{MAE}} & \textbf{MSE} & \multicolumn{1}{c|}{\textbf{MAE}} & \textbf{MSE} & \multicolumn{1}{c|}{\textbf{MAE}} & \textbf{MSE} & \multicolumn{1}{c|}{\textbf{MAE}} & \textbf{MSE} & \multicolumn{1}{c|}{\textbf{MAE}} & \textbf{MSE} & \multicolumn{1}{c|}{\textbf{MAE}} & \textbf{MSE} & \multicolumn{1}{c|}{\textbf{MAE}} & \textbf{MSE} & \textbf{MAE} & \textbf{MSE} & \multicolumn{1}{c|}{\textbf{MAE}} & \textbf{MSE} & \textbf{MAE} \\ \midrule

\multicolumn{1}{c|}{} & \textbf{96} & {\color[HTML]{FF0000} \textbf{0.315}} & \multicolumn{1}{c|}{{\color[HTML]{FF0000} \textbf{0.352}}} & 
0.321 & 0.360 & 0.320 & \multicolumn{1}{c|}{{\color[HTML]{5B9BD5} \textbf{0.357}}} &
0.342 & \multicolumn{1}{c|}{0.371} & {\color[HTML]{5B9BD5} \textbf{0.319}} & 0.359 & 0.334 & \multicolumn{1}{c|}{0.368} & {\color[HTML]{5B9BD5} \textbf{0.319}} & \multicolumn{1}{c|}{0.366} & 0.335 & \multicolumn{1}{c|}{0.373} & 0.329 & \multicolumn{1}{c|}{0.367} & 0.404 & \multicolumn{1}{c|}{0.426} & 0.338 & \multicolumn{1}{c|}{0.375} & 0.345 & \multicolumn{1}{c|}{0.372} & 0.418 & \multicolumn{1}{c|}{0.438} & 0.379 & \multicolumn{1}{c|}{0.419} & 0.386 & \multicolumn{1}{c|}{0.398} & 0.505 & \multicolumn{1}{c|}{0.475} & 0.379 & 0.446 \\

\multicolumn{1}{c|}{} & \textbf{192} & {\color[HTML]{FF0000} \textbf{0.351}} & \multicolumn{1}{c|}{{\color[HTML]{FF0000} \textbf{0.379}}} &
0.363 & 0.382 & {\color[HTML]{5B9BD5} \textbf{0.361}} & {\color[HTML]{5B9BD5} \textbf{0.381}} &
0.384 & \multicolumn{1}{c|}{0.390} & 0.364 & 0.382 & 0.377 & \multicolumn{1}{c|}{0.391} & 0.376 & \multicolumn{1}{c|}{0.397} & 0.372 & \multicolumn{1}{c|}{0.390} & 0.367 & \multicolumn{1}{c|}{0.385} & 0.450 & \multicolumn{1}{c|}{0.451} & 0.374 & \multicolumn{1}{c|}{0.387} & 0.380 & \multicolumn{1}{c|}{0.389} & 0.439 & \multicolumn{1}{c|}{0.450} & 0.426 & \multicolumn{1}{c|}{0.441} & 0.459 & \multicolumn{1}{c|}{0.444} & 0.553 & \multicolumn{1}{c|}{0.496} & 0.470 & 0.428 \\

\multicolumn{1}{c|}{} & \textbf{336} & {\color[HTML]{FF0000} \textbf{0.387}} & \multicolumn{1}{c|}{{\color[HTML]{5B9BD5} \textbf{0.404}}} & 
0.395 & \multicolumn{1}{c|}{{\color[HTML]{FF0000} \textbf{0.403}}} & \multicolumn{1}{c}{{\color[HTML]{5B9BD5} \textbf{0.390}}} & {\color[HTML]{5B9BD5} \textbf{0.404}} &
0.424 & \multicolumn{1}{c|}{0.414} & 0.399 & 0.405 & 0.426 & \multicolumn{1}{c|}{0.420} & 0.417 & \multicolumn{1}{c|}{0.422} & 0.403 & \multicolumn{1}{c|}{0.411} & 0.399 & \multicolumn{1}{c|}{0.410} & 0.532 & \multicolumn{1}{c|}{0.515} & 0.410 & \multicolumn{1}{c|}{0.411} & 0.413 & \multicolumn{1}{c|}{0.413} & 0.490 & \multicolumn{1}{c|}{0.485} & 0.445 & \multicolumn{1}{c|}{0.459} & 0.495 & \multicolumn{1}{c|}{0.464} & 0.621 & \multicolumn{1}{c|}{0.537} & 0.473 & 0.430 \\

\multicolumn{1}{c|}{} & \textbf{720} & {\color[HTML]{FF0000} \textbf{0.442}} & \multicolumn{1}{c|}{{\color[HTML]{FF0000} \textbf{0.429}}} & 
0.462 & {\color[HTML]{5B9BD5} \textbf{0.438}} & {\color[HTML]{5B9BD5} \textbf{0.454}} & 0.441 &
0.483 & \multicolumn{1}{c|}{0.445} & 0.467 & \multicolumn{1}{c|}{0.442} & 0.491 & \multicolumn{1}{c|}{0.459} & 0.481 & \multicolumn{1}{c|}{0.458} & 0.461 & \multicolumn{1}{c|}{0.442} & {\color[HTML]{5B9BD5} \textbf{0.454}} & 0.439 & 0.666 & \multicolumn{1}{c|}{0.589} & 0.478 & \multicolumn{1}{c|}{0.450} & 0.474 & \multicolumn{1}{c|}{0.453} & 0.595 & \multicolumn{1}{c|}{0.550} & 0.543 & \multicolumn{1}{c|}{0.490} & 0.585 & \multicolumn{1}{c|}{0.516} & 0.671 & \multicolumn{1}{c|}{0.561} & 0.553 & 0.479 \\ \cmidrule(l){2-36} 

\multicolumn{1}{c|}{\multirow{-5}{*}{\textbf{ETTm1}}} & \textbf{Avg} & {\color[HTML]{FF0000} \textbf{0.373}} & \multicolumn{1}{c|}{{\color[HTML]{FF0000} \textbf{0.390}}} &
0.385 & 0.396 & {\color[HTML]{5B9BD5} \textbf{0.381}} & {\color[HTML]{5B9BD5} \textbf{0.395}} &
0.408 & \multicolumn{1}{c|}{0.405} & 0.387& \multicolumn{1}{c|}{0.397} & 0.407 & \multicolumn{1}{c|}{0.410} & 0.398 & \multicolumn{1}{c|}{0.411} & 0.393 & \multicolumn{1}{c|}{0.404} & 0.387 & \multicolumn{1}{c|}{0.400} & 0.513 & \multicolumn{1}{c|}{0.495} & 0.400 & \multicolumn{1}{c|}{0.406} & 0.403 & \multicolumn{1}{c|}{0.407} & 0.486 & \multicolumn{1}{c|}{0.481} & 0.448 & \multicolumn{1}{c|}{0.452} & 0.481 & \multicolumn{1}{c|}{0.456} & 0.588 & \multicolumn{1}{c|}{0.517} & 0.469 & 0.446 \\ \midrule

\multicolumn{1}{c|}{} & \textbf{96} & 0.175 & 0.257
& {\color[HTML]{5B9BD5} \textbf{0.172}} & {\color[HTML]{FF0000} \textbf{0.255}} & 0.175 & 0.258
& {\color[HTML]{FF0000} \textbf{0.170}} & \multicolumn{1}{c|}{{\color[HTML]{5B9BD5} \textbf{0.256}}} & 0.180 & \multicolumn{1}{c|}{0.266} & 0.180 & \multicolumn{1}{c|}{0.264} & 0.177 & \multicolumn{1}{c|}{0.262} & 0.176 & \multicolumn{1}{c|}{0.266} & 0.175 & \multicolumn{1}{c|}{0.259} & 0.287 & \multicolumn{1}{c|}{0.366} & 0.187 & \multicolumn{1}{c|}{0.267} & 0.193 & \multicolumn{1}{c|}{0.292} & 0.286 & \multicolumn{1}{c|}{0.377} & 0.203 & \multicolumn{1}{c|}{0.287} & 0.192 & \multicolumn{1}{c|}{0.274} & 0.255 & \multicolumn{1}{c|}{0.339} & 0.203 & 0.299 \\

\multicolumn{1}{c|}{} & \textbf{192} & {\color[HTML]{FF0000} \textbf{0.233}} & \multicolumn{1}{c|}{{\color[HTML]{FF0000} \textbf{0.296}}} 
& {\color[HTML]{5B9BD5} \textbf{0.237}} & {\color[HTML]{5B9BD5} \textbf{0.299}} & {\color[HTML]{5B9BD5} \textbf{0.237}} & {\color[HTML]{5B9BD5} \textbf{0.299}}
& 0.247 & \multicolumn{1}{c|}{0.303} & 0.245 & \multicolumn{1}{c|}{0.306} & 0.250 & \multicolumn{1}{c|}{0.309} & 0.247 & \multicolumn{1}{c|}{0.307} & 0.240 & \multicolumn{1}{c|}{0.307} & 0.241 & \multicolumn{1}{c|}{0.302} & 0.414 & \multicolumn{1}{c|}{0.492} & 0.249 & \multicolumn{1}{c|}{0.309} & 0.284 & \multicolumn{1}{c|}{0.362} & 0.399 & \multicolumn{1}{c|}{0.445} & 0.269 & \multicolumn{1}{c|}{0.328} & 0.280 & \multicolumn{1}{c|}{0.339} & 0.281 & \multicolumn{1}{c|}{0.340} & 0.265 & 0.328 \\

\multicolumn{1}{c|}{} & \textbf{336} & {\color[HTML]{FF0000} \textbf{0.291}} & \multicolumn{1}{c|}{{\color[HTML]{FF0000} \textbf{0.334}}} 
& 0.299 & 0.398 &  {\color[HTML]{5B9BD5} \textbf{0.298}} &  {\color[HTML]{5B9BD5} \textbf{0.340}}
& 0.309 & \multicolumn{1}{c|}{0.343} & 0.307 & \multicolumn{1}{c|}{0.345} & 0.311 & \multicolumn{1}{c|}{0.348} & 0.312 & \multicolumn{1}{c|}{0.346} & 0.304 & \multicolumn{1}{c|}{0.345} & 0.305 & 0.343& 0.597 & \multicolumn{1}{c|}{0.542} & 0.321 & \multicolumn{1}{c|}{0.351} & 0.369 & \multicolumn{1}{c|}{0.427} & 0.637 & \multicolumn{1}{c|}{0.591} & 0.325 & \multicolumn{1}{c|}{0.366} & 0.334 & \multicolumn{1}{c|}{0.361} & 0.339 & \multicolumn{1}{c|}{0.372} & 0.365 & 0.374 \\

\multicolumn{1}{c|}{} & \textbf{720} & {\color[HTML]{FF0000} \textbf{0.389}} & \multicolumn{1}{c|}{{\color[HTML]{FF0000} \textbf{0.390}}} 
& 0.397 & {\color[HTML]{5B9BD5} \textbf{0.394}} & {\color[HTML]{5B9BD5} \textbf{0.391}} & 0.396
& 0.411 & \multicolumn{1}{c|}{0.403} & 0.406 & \multicolumn{1}{c|}{0.402} & 0.412 & \multicolumn{1}{c|}{0.407} & 0.414 & \multicolumn{1}{c|}{0.403} & 0.406 & 0.400 & 0.402 & \multicolumn{1}{c|}{0.400} & 1.730 & \multicolumn{1}{c|}{1.042} & 0.408 & \multicolumn{1}{c|}{0.403} & 0.554 & \multicolumn{1}{c|}{0.522} & 0.960 & \multicolumn{1}{c|}{0.735} & 0.421 & \multicolumn{1}{c|}{0.415} & 0.417 & \multicolumn{1}{c|}{0.413} & 0.433 & \multicolumn{1}{c|}{0.432} & 0.461 & 0.459 \\ \cmidrule(l){2-36} 

\multicolumn{1}{c|}{\multirow{-5}{*}{\textbf{ETTm2}}} & \textbf{Avg} & {\color[HTML]{FF0000} \textbf{0.272}} & \multicolumn{1}{c|}{{\color[HTML]{FF0000} \textbf{0.319}}} 
& 0.276 & {\color[HTML]{5B9BD5} \textbf{0.321}} & {\color[HTML]{5B9BD5} \textbf{0.275}} & 0.323
& 0.284 & \multicolumn{1}{c|}{0.326} & 0.285 & \multicolumn{1}{c|}{0.330} & 0.288 & \multicolumn{1}{c|}{0.332} & 0.288 & \multicolumn{1}{c|}{0.330} & 0.282 & \multicolumn{1}{c|}{0.330} & 0.281 & \multicolumn{1}{c|}{0.326} & 0.757 & \multicolumn{1}{c|}{0.611} & 0.291 & \multicolumn{1}{c|}{0.333} & 0.350 & \multicolumn{1}{c|}{0.401} & 0.571 & \multicolumn{1}{c|}{0.537} & 0.305 & \multicolumn{1}{c|}{0.349} & 0.306 & \multicolumn{1}{c|}{0.347} & 0.327 & \multicolumn{1}{c|}{0.371} & 0.324 & 0.365 \\ \midrule

\multicolumn{1}{c|}{} & \textbf{96} & {\color[HTML]{FF0000} \textbf{0.372}} & \multicolumn{1}{c|}{{\color[HTML]{FF0000} \textbf{0.391}}} & 
{\color[HTML]{5B9BD5} \textbf{0.374}} & {\color[HTML]{FF0000} \textbf{0.391}} & 0.375 & 0.400 & 
0.385 & \multicolumn{1}{c|}{{\color[HTML]{5B9BD5} \textbf{0.393}}} & 0.376 & \multicolumn{1}{c|}{0.397} & 0.386 & \multicolumn{1}{c|}{0.405} & 0.390 & \multicolumn{1}{c|}{0.411} & 0.382 & \multicolumn{1}{c|}{0.398} & 0.414 & \multicolumn{1}{c|}{0.419} & 0.423 & \multicolumn{1}{c|}{0.448} & 0.384 & \multicolumn{1}{c|}{0.402} & 0.386 & \multicolumn{1}{c|}{0.400} & 0.654 & \multicolumn{1}{c|}{0.599} & 0.3764 & \multicolumn{1}{c|}{0.419} & 0.513 & \multicolumn{1}{c|}{0.491} & 0.449 & \multicolumn{1}{c|}{0.459} & 0.515 & 0.517 \\

\multicolumn{1}{c|}{} & \textbf{192} & 0.425 & \multicolumn{1}{c|}{{\color[HTML]{FF0000} \textbf{0.420}}} & 
{\color[HTML]{5B9BD5} \textbf{0.424}} & {\color[HTML]{5B9BD5} \textbf{0.421}} & 0.429 & {\color[HTML]{5B9BD5} \textbf{0.421}}
& {\color[HTML]{5B9BD5} \textbf{0.424}} & \multicolumn{1}{c|}{0.438} & 0.431 & \multicolumn{1}{c|}{0.427} & 0.441 & \multicolumn{1}{c|}{0.436} & 0.442 & \multicolumn{1}{c|}{0.442} & 0.427 & 0.425 & 0.460 & \multicolumn{1}{c|}{0.445} & 0.471 & \multicolumn{1}{c|}{0.474} & 0.436 & \multicolumn{1}{c|}{0.429} & 0.437 & \multicolumn{1}{c|}{0.432} & 0.719 & \multicolumn{1}{c|}{0.631} & {\color[HTML]{FF0000} \textbf{0.420}} & \multicolumn{1}{c|}{0.448} & 0.534 & \multicolumn{1}{c|}{0.504} & 0.500 & \multicolumn{1}{c|}{0.482} & 0.553 & 0.522 \\

\multicolumn{1}{c|}{} & \textbf{336} & {\color[HTML]{FF0000} \textbf{0.457}} & \multicolumn{1}{c|}{{\color[HTML]{FF0000} \textbf{0.438}}} &
0.464 & {\color[HTML]{5B9BD5} \textbf{0.441}} & 0.484 & 0.458 &
0.482 & \multicolumn{1}{c|}{0.447} & 0.473 & \multicolumn{1}{c|}{0.449} & 0.487 & \multicolumn{1}{c|}{0.458} & 0.480 & \multicolumn{1}{c|}{0.468} & 0.465 & \multicolumn{1}{c|}{0.445} & 0.501 & \multicolumn{1}{c|}{0.466} & 0.570 & \multicolumn{1}{c|}{0.546} & 0.491 & \multicolumn{1}{c|}{0.469} & 0.481 & \multicolumn{1}{c|}{0.459} & 0.778 & \multicolumn{1}{c|}{0.659} & {\color[HTML]{5B9BD5} \textbf{0.459}} & \multicolumn{1}{c|}{0.465} & 0.588 & \multicolumn{1}{c|}{0.535} & 0.521 & \multicolumn{1}{c|}{0.496} & 0.612 & 0.577 \\

\multicolumn{1}{c|}{} & \textbf{720} & {\color[HTML]{FF0000} \textbf{0.468}} & \multicolumn{1}{c|}{{\color[HTML]{FF0000} \textbf{0.462}}} & 
{\color[HTML]{5B9BD5} \textbf{0.470}} & {\color[HTML]{5B9BD5} \textbf{0.466}} & 0.498 & 0.482 &
0.489 & \multicolumn{1}{c|}{0.473} & 0.476 & \multicolumn{1}{c|}{0.474} & 0.503 & \multicolumn{1}{c|}{0.491} & 0.494 & \multicolumn{1}{c|}{0.488} & 0.472 & \multicolumn{1}{c|}{0.468} & 0.500 & \multicolumn{1}{c|}{0.488} & 0.653 & \multicolumn{1}{c|}{0.621} & 0.521 & \multicolumn{1}{c|}{0.500} & 0.519 & \multicolumn{1}{c|}{0.516} & 0.836 & \multicolumn{1}{c|}{0.699} & 0.506 & \multicolumn{1}{c|}{0.507} & 0.643 & \multicolumn{1}{c|}{0.616} & 0.514 & \multicolumn{1}{c|}{0.512} & 0.609 & 0.597 \\ \cmidrule(l){2-36} 

\multicolumn{1}{c|}{\multirow{-5}{*}{\textbf{ETTh1}}} & \textbf{Avg} & {\color[HTML]{FF0000} \textbf{0.431}} & \multicolumn{1}{c|}{{\color[HTML]{FF0000} \textbf{0.428}}} &
{\color[HTML]{5B9BD5} \textbf{0.433}} & {\color[HTML]{5B9BD5} \textbf{0.430}} & 0.447 & 0.440
& 0.445 & \multicolumn{1}{c|}{0.438} & 0.439 & \multicolumn{1}{c|}{0.437} & 0.454 & \multicolumn{1}{c|}{0.448} & 0.452 & \multicolumn{1}{c|}{0.452} & 0.437 & \multicolumn{1}{c|}{0.434} & 0.469 & \multicolumn{1}{c|}{0.455} & 0.529 & \multicolumn{1}{c|}{0.522} & 0.458 & \multicolumn{1}{c|}{0.450} & 0.456 & \multicolumn{1}{c|}{0.452} & 0.747 & \multicolumn{1}{c|}{0.647} & 0.440 & \multicolumn{1}{c|}{0.460} & 0.570 & \multicolumn{1}{c|}{0.537} & 0.496 & \multicolumn{1}{c|}{0.487} & 0.572 & 0.553 \\ \midrule

\multicolumn{1}{c|}{} & \textbf{96} & 0.292 & \multicolumn{1}{c|}{{\color[HTML]{5B9BD5} \textbf{0.340}}} 
& {\color[HTML]{5B9BD5} \textbf{0.290}} & {\color[HTML]{FF0000} \textbf{0.338}} & {\color[HTML]{FF0000} \textbf{0.289}} & 0.341
& 0.292 & \multicolumn{1}{c|}{0.341} & 0.292 & \multicolumn{1}{c|}{0.344} & 0.297 & \multicolumn{1}{c|}{0.349} & 0.328 & \multicolumn{1}{c|}{0.371} & 0.309 & \multicolumn{1}{c|}{0.359} & 0.302 & \multicolumn{1}{c|}{0.348} & 0.745 & \multicolumn{1}{c|}{0.584} & 0.340 & \multicolumn{1}{c|}{0.374} & 0.333 & \multicolumn{1}{c|}{0.387} & 0.707 & \multicolumn{1}{c|}{0.621} & 0.358 & \multicolumn{1}{c|}{0.397} & 0.476 & \multicolumn{1}{c|}{0.458} & 0.346 & \multicolumn{1}{c|}{0.388} & 0.354 & 0.454 \\

\multicolumn{1}{c|}{} & \textbf{192} & 0.375 & \multicolumn{1}{c|}{{\color[HTML]{FF0000} \textbf{0.390}}} 
& {\color[HTML]{5B9BD5} \textbf{0.374}} & {\color[HTML]{FF0000} \textbf{0.390}} & {\color[HTML]{FF0000} \textbf{0.372}} & {\color[HTML]{5B9BD5} \textbf{0.392}}
& 0.378 & \multicolumn{1}{c|}{{\color[HTML]{5B9BD5} \textbf{0.392}}} & 0.377 & \multicolumn{1}{c|}{0.396} & 0.380 & \multicolumn{1}{c|}{0.400} & 0.402 & \multicolumn{1}{c|}{0.414} & 0.390 & \multicolumn{1}{c|}{0.406} & 0.388 & \multicolumn{1}{c|}{0.400} & 0.877 & \multicolumn{1}{c|}{0.656} & 0.402 & \multicolumn{1}{c|}{0.414} & 0.477 & \multicolumn{1}{c|}{0.476} & 0.860 & \multicolumn{1}{c|}{0.689} & 0.429 & \multicolumn{1}{c|}{0.439} & 0.512 & \multicolumn{1}{c|}{0.493} & 0.456 & \multicolumn{1}{c|}{0.452} & 0.457 & 0.464 \\

\multicolumn{1}{c|}{} & \textbf{336} & {\color[HTML]{5B9BD5} \textbf{0.400}} & 0.424
& 0.406 & \multicolumn{1}{c|}{{\color[HTML]{5B9BD5} \textbf{0.420}}} & {\color[HTML]{FF0000} \textbf{0.386}} & {\color[HTML]{FF0000} \textbf{0.414}}
& 0.418 & 0.427 & 0.417 & \multicolumn{1}{c|}{0.430} & 0.428 & \multicolumn{1}{c|}{0.432} & 0.435 & \multicolumn{1}{c|}{0.443} & 0.426 & \multicolumn{1}{c|}{0.444} & 0.426 & \multicolumn{1}{c|}{0.433} & 1.043 & \multicolumn{1}{c|}{0.731} & 0.452 & \multicolumn{1}{c|}{0.452} & 0.594 & \multicolumn{1}{c|}{0.541} & 1.000 & \multicolumn{1}{c|}{0.744} & 0.496 & \multicolumn{1}{c|}{0.487} & 0.552 & \multicolumn{1}{c|}{0.551} & 0.482 & \multicolumn{1}{c|}{0.486} & 0.515 & 0.540 \\

\multicolumn{1}{c|}{} & \textbf{720} & 0.427 & \multicolumn{1}{c|}{0.441} 
& {\color[HTML]{5B9BD5} \textbf{0.418}} & {\color[HTML]{5B9BD5} \textbf{0.437}} & {\color[HTML]{FF0000} \textbf{0.412}} & {\color[HTML]{FF0000} \textbf{0.434}}
& 0.423 & \multicolumn{1}{c|}{0.441} & 0.423 & \multicolumn{1}{c|}{0.443} & 0.427 & \multicolumn{1}{c|}{0.445} & 0.417 & \multicolumn{1}{c|}{0.441} & 0.445 & \multicolumn{1}{c|}{0.464} & 0.431 & \multicolumn{1}{c|}{0.446} & 1.104 & \multicolumn{1}{c|}{0.763} & 0.462 & \multicolumn{1}{c|}{0.468} & 0.831 & \multicolumn{1}{c|}{0.657} & 1.249 & \multicolumn{1}{c|}{0.838} & 0.463 & \multicolumn{1}{c|}{0.474} & 0.562 & \multicolumn{1}{c|}{0.560} & 0.515 & \multicolumn{1}{c|}{0.511} & 0.532 & 0.576 \\ \cmidrule(l){2-36} 

\multicolumn{1}{c|}{\multirow{-5}{*}{\textbf{ETTh2}}} & \textbf{Avg} & 0.374 & 0.398 
& {\color[HTML]{5B9BD5} \textbf{0.372}} & {\color[HTML]{5B9BD5} \textbf{0.396}} & {\color[HTML]{FF0000} \textbf{0.364}} & {\color[HTML]{FF0000} \textbf{0.395}}
& 0.378 & \multicolumn{1}{c|}{0.399} & 0.377 & 0.403 & 0.383 & \multicolumn{1}{c|}{0.407} & 0.396 & \multicolumn{1}{c|}{0.417} & 0.393 & \multicolumn{1}{c|}{0.418} & 0.387 & \multicolumn{1}{c|}{0.407} & 0.942 & \multicolumn{1}{c|}{0.684} & 0.414 & \multicolumn{1}{c|}{0.427} & 0.559 & \multicolumn{1}{c|}{0.515} & 0.954 & \multicolumn{1}{c|}{0.723} & 0.437 & \multicolumn{1}{c|}{0.449} & 0.526 & \multicolumn{1}{c|}{0.516} & 0.450 & \multicolumn{1}{c|}{0.459} & 0.465 & 0.509 \\ \midrule

\multicolumn{1}{c|}{} & \textbf{96} & {\color[HTML]{FF0000} \textbf{0.136}} & \multicolumn{1}{c|}{{\color[HTML]{FF0000} \textbf{0.238}}} 
& 0.151 & 0.245 & 0.153 & 0.247
& 0.160 & \multicolumn{1}{c|}{0.250} & 0.150 & \multicolumn{1}{c|}{0.242} & {\color[HTML]{5B9BD5} \textbf{0.148}} & \multicolumn{1}{c|}{{\color[HTML]{5B9BD5} \textbf{0.240}}} & 0.165 & \multicolumn{1}{c|}{0.274} & 0.173 & \multicolumn{1}{c|}{0.275} & 0.181 & \multicolumn{1}{c|}{0.270} & 0.219 & \multicolumn{1}{c|}{0.314} & 0.168 & \multicolumn{1}{c|}{0.272} & 0.197 & \multicolumn{1}{c|}{0.282} & 0.247 & \multicolumn{1}{c|}{0.345} & 0.193 & \multicolumn{1}{c|}{0.308} & 0.169 & \multicolumn{1}{c|}{0.273} & 0.201 & \multicolumn{1}{c|}{0.317} & 0.217 & 0.318 \\

\multicolumn{1}{c|}{} & \textbf{192} & {\color[HTML]{FF0000} \textbf{0.152}} & \multicolumn{1}{c|}{{\color[HTML]{FF0000} \textbf{0.241}}} 
& 0.163 & 0.256 & 0.166 & 0.256
& 0.183 & \multicolumn{1}{c|}{0.267} & 0.167 & \multicolumn{1}{c|}{0.255} & {\color[HTML]{5B9BD5} \textbf{0.162}} & \multicolumn{1}{c|}{{\color[HTML]{5B9BD5} \textbf{0.253}}} & 0.184 & \multicolumn{1}{c|}{0.292} & 0.195 & \multicolumn{1}{c|}{0.288} & 0.188 & \multicolumn{1}{c|}{0.274} & 0.231 & \multicolumn{1}{c|}{0.322} & 0.184 & \multicolumn{1}{c|}{0.289} & 0.196 & \multicolumn{1}{c|}{0.285} & 0.257 & \multicolumn{1}{c|}{0.355} & 0.201 & \multicolumn{1}{c|}{0.315} & 0.182 & \multicolumn{1}{c|}{0.286} & 0.222 & \multicolumn{1}{c|}{0.334} & 0.238 & 0.352 \\

\multicolumn{1}{c|}{} & \textbf{336} & {\color[HTML]{FF0000} \textbf{0.169}} & \multicolumn{1}{c|}{{\color[HTML]{5B9BD5} \textbf{0.272}}} 
& 0.180 & 0.274 & 0.185 & 0.277
& 0.200 & \multicolumn{1}{c|}{0.284} & 0.182 & \multicolumn{1}{c|}{0.270} & {\color[HTML]{5B9BD5} \textbf{0.178}} & \multicolumn{1}{c|}{{\color[HTML]{FF0000} \textbf{0.269}}} & 0.195 & \multicolumn{1}{c|}{0.302} & 0.206 & \multicolumn{1}{c|}{0.300} & 0.204 & \multicolumn{1}{c|}{0.293} & 0.246 & \multicolumn{1}{c|}{0.337} & 0.198 & \multicolumn{1}{c|}{0.300} & 0.209 & \multicolumn{1}{c|}{0.301} & 0.269 & \multicolumn{1}{c|}{0.369} & 0.214 & \multicolumn{1}{c|}{0.329} & 0.200 & \multicolumn{1}{c|}{0.304} & 0.231 & \multicolumn{1}{c|}{0.338} & 0.260 & 0.348 \\

\multicolumn{1}{c|}{} & \textbf{720} & 0.223 & \multicolumn{1}{c|}{{\color[HTML]{FF0000} \textbf{0.298}}} 
& 0.224 & 0.311 & 0.225 & 0.310
& 0.230 & \multicolumn{1}{c|}{0.310} & {\color[HTML]{5B9BD5} \textbf{0.221}} & \multicolumn{1}{c|}{{\color[HTML]{5B9BD5} \textbf{0.302}}} & 0.225 & \multicolumn{1}{c|}{0.317} & 0.231 & \multicolumn{1}{c|}{0.332} & 0.231 & \multicolumn{1}{c|}{0.335} & 0.246 & \multicolumn{1}{c|}{0.324} & 0.280 & \multicolumn{1}{c|}{0.363} & {\color[HTML]{FF0000} \textbf{0.220}} & \multicolumn{1}{c|}{0.320} & 0.245 & \multicolumn{1}{c|}{0.333} & 0.299 & \multicolumn{1}{c|}{0.390} & 0.246 & \multicolumn{1}{c|}{0.355} & 0.222 & \multicolumn{1}{c|}{0.321} & 0.254 & \multicolumn{1}{c|}{0.361} & 0.290 & 0.369 \\ \cmidrule(l){2-36} 

\multicolumn{1}{c|}{\multirow{-5}{*}{\textbf{ECL}}} & \textbf{Avg} & {\color[HTML]{FF0000} \textbf{0.170}} & \multicolumn{1}{c|}{{\color[HTML]{FF0000} \textbf{0.262}}} 
& 0.180 & 0.271 & 0.182 & 0.272
& 0.193 & \multicolumn{1}{c|}{0.278} & 0.180 & \multicolumn{1}{c|}{{\color[HTML]{5B9BD5} \textbf{0.267}}} & {\color[HTML]{5B9BD5} \textbf{0.178}} & \multicolumn{1}{c|}{0.270} & 0.194 & \multicolumn{1}{c|}{0.300} & 0.201 & \multicolumn{1}{c|}{0.300} & 0.205 & \multicolumn{1}{c|}{0.290} & 0.244 & \multicolumn{1}{c|}{0.334} & 0.193 & \multicolumn{1}{c|}{0.295} & 0.212 & \multicolumn{1}{c|}{0.300} & 0.268 & \multicolumn{1}{c|}{0.365} & 0.214 & \multicolumn{1}{c|}{0.327} & 0.193 & \multicolumn{1}{c|}{0.296} & 0.227 & \multicolumn{1}{c|}{0.338} & 0.251 & 0.347 \\ \midrule

\multicolumn{1}{c|}{} & \textbf{96} & {\color[HTML]{5B9BD5} \textbf{0.083}} & \multicolumn{1}{c|}{0.207} 
& {\color[HTML]{FF0000} \textbf{0.081}} & \multicolumn{1}{c|}{{\color[HTML]{5B9BD5} \textbf{0.199}}} & 0.085$^*$ & 0.204$^*$
& {\color[HTML]{FF0000} \textbf{0.081}} & \multicolumn{1}{c|}{{\color[HTML]{FF0000} \textbf{0.198}}} & 0.085 & \multicolumn{1}{c|}{0.205} & 0.086 & \multicolumn{1}{c|}{0.206} & 0.102 & \multicolumn{1}{c|}{0.230} & 0.084 & 0.203 & 0.088 & \multicolumn{1}{c|}{0.205} & 0.256 & \multicolumn{1}{c|}{0.367} & 0.107 & \multicolumn{1}{c|}{0.234} & 0.088 & \multicolumn{1}{c|}{0.218} & 0.267 & \multicolumn{1}{c|}{0.396} & 0.148 & \multicolumn{1}{c|}{0.278} & 0.111 & \multicolumn{1}{c|}{0.237} & 0.197 & \multicolumn{1}{c|}{0.323} & 0.102 & 0.228 \\

\multicolumn{1}{c|}{} & \textbf{192} & {\color[HTML]{FF0000} \textbf{0.169}} & \multicolumn{1}{c|}{{\color[HTML]{FF0000} \textbf{0.290}}} 
& {\color[HTML]{5B9BD5} \textbf{0.171}} & {\color[HTML]{5B9BD5} \textbf{0.294}} & 0.172$^*$ & 0.295$^*$
& 0.174 & \multicolumn{1}{c|}{0.297} & 0.176 & \multicolumn{1}{c|}{0.299} & 0.177 & \multicolumn{1}{c|}{0.299} & 0.195 & \multicolumn{1}{c|}{0.317} & {\color[HTML]{5B9BD5} \textbf{0.171}} & \multicolumn{1}{c|}{{\color[HTML]{5B9BD5} \textbf{0.294}}} & 0.176 & \multicolumn{1}{c|}{0.299} & 0.470 & \multicolumn{1}{c|}{0.509} & 0.226 & \multicolumn{1}{c|}{0.344} & 0.176 & \multicolumn{1}{c|}{0.315} & 0.351 & \multicolumn{1}{c|}{0.459} & 0.271 & \multicolumn{1}{c|}{0.315} & 0.219 & \multicolumn{1}{c|}{0.335} & 0.300 & \multicolumn{1}{c|}{0.369} & 0.267 & 0.335 \\

\multicolumn{1}{c|}{} & \textbf{336} & 0.317 & \multicolumn{1}{c|}{{\color[HTML]{5B9BD5} \textbf{0.402}}} 
& 0.321 & 0.409 & 0.377$^*$ & 0.445$^*$
& 0.322 & \multicolumn{1}{c|}{0.409} & 0.328 & \multicolumn{1}{c|}{0.415} & 0.331 & \multicolumn{1}{c|}{0.417} & 0.359 & \multicolumn{1}{c|}{0.436} & 0.319 & \multicolumn{1}{c|}{0.407} & {\color[HTML]{FF0000} \textbf{0.301}} & \multicolumn{1}{c|}{{\color[HTML]{FF0000} \textbf{0.397}}} & 1.268 & \multicolumn{1}{c|}{0.883} & 0.367 & \multicolumn{1}{c|}{0.448} & {\color[HTML]{5B9BD5} \textbf{0.313}} & \multicolumn{1}{c|}{0.427} & 1.324 & \multicolumn{1}{c|}{0.853} & 0.460 & \multicolumn{1}{c|}{0.427} & 0.421 & \multicolumn{1}{c|}{0.476} & 0.509 & \multicolumn{1}{c|}{0.524} & 0.393 & 0.457 \\

\multicolumn{1}{c|}{} & \textbf{720} & {\color[HTML]{5B9BD5} \textbf{0.811}} & \multicolumn{1}{c|}{{\color[HTML]{FF0000} \textbf{0.671}}} 
& 0.837 & 0.688 & 0.934$^*$ & 0.728$^*$
& 0.830 & \multicolumn{1}{c|}{0.686} & 0.830 & \multicolumn{1}{c|}{0.688} & 0.847 & \multicolumn{1}{c|}{0.691} & 0.940 & \multicolumn{1}{c|}{0.738} & {\color[HTML]{FF0000} \textbf{0.805}} & \multicolumn{1}{c|}{{\color[HTML]{5B9BD5} \textbf{0.677}}} & 0.901 & \multicolumn{1}{c|}{0.714} & 1.767 & \multicolumn{1}{c|}{1.068} & 0.964 & \multicolumn{1}{c|}{0.746} & 0.839 & \multicolumn{1}{c|}{0.695} & 1.058 & \multicolumn{1}{c|}{0.797} & 1.195 & \multicolumn{1}{c|}{0.695} & 1.092 & \multicolumn{1}{c|}{0.769} & 1.447 & \multicolumn{1}{c|}{0.941} & 1.090 & 0.811 \\ \cmidrule(l){2-36} 

\multicolumn{1}{c|}{\multirow{-5}{*}{\textbf{Exchange}}} & \textbf{Avg} & {\color[HTML]{FF0000} \textbf{0.344}} & \multicolumn{1}{c|}{{\color[HTML]{FF0000} \textbf{0.393}}} 
& 0.352 & 0.397 & 0.392$^*$ & 0.418$^*$
& 0.352 & \multicolumn{1}{c|}{0.398} & 0.355 & \multicolumn{1}{c|}{0.402} & 0.360 & \multicolumn{1}{c|}{0.403} & 0.399 & \multicolumn{1}{c|}{0.430} & {\color[HTML]{5B9BD5} \textbf{0.345}} & \multicolumn{1}{c|}{{\color[HTML]{5B9BD5} \textbf{0.395}}} & 0.367 & \multicolumn{1}{c|}{0.404} & 0.940 & \multicolumn{1}{c|}{0.707} & 0.416 & \multicolumn{1}{c|}{0.443} & 0.354 & \multicolumn{1}{c|}{0.414} & 0.750 & \multicolumn{1}{c|}{0.626} & 0.519 & \multicolumn{1}{c|}{0.429} & 0.461 & \multicolumn{1}{c|}{0.454} & 0.613 & \multicolumn{1}{c|}{0.539} & 0.463 & 0.458 \\ \midrule

\multicolumn{1}{c|}{} & \textbf{96} & {\color[HTML]{FF0000} \textbf{0.382}} & \multicolumn{1}{c|}{{\color[HTML]{FF0000} \textbf{0.257}}} 
& 0.448 & 0.309 & 0.462 & 0.285
& 0.501 & \multicolumn{1}{c|}{0.326} & 0.454 & \multicolumn{1}{c|}{0.310} & {\color[HTML]{5B9BD5} \textbf{0.395}} & \multicolumn{1}{c|}{{\color[HTML]{5B9BD5} \textbf{0.268}}} & 0.605$^*$ & \multicolumn{1}{c|}{0.344$^*$} & 0.570 & \multicolumn{1}{c|}{0.310} & 0.462 & \multicolumn{1}{c|}{0.295} & 0.522 & \multicolumn{1}{c|}{0.290} & 0.593 & \multicolumn{1}{c|}{0.321} & 0.650 & \multicolumn{1}{c|}{0.396} & 0.788 & \multicolumn{1}{c|}{0.499} & 0.587 & \multicolumn{1}{c|}{0.366} & 0.612 & \multicolumn{1}{c|}{0.338} & 0.613 & \multicolumn{1}{c|}{0.388} & 0.660 & 0.437 \\

\multicolumn{1}{c|}{} & \textbf{192} & {\color[HTML]{FF0000} \textbf{0.411}} & \multicolumn{1}{c|}{{\color[HTML]{FF0000} \textbf{0.269}}} 
& 0.455 & 0.307 & 0.473 & 0.296
& 0.499 & \multicolumn{1}{c|}{0.321} & 0.468 & \multicolumn{1}{c|}{0.315} & {\color[HTML]{5B9BD5} \textbf{0.417}} & \multicolumn{1}{c|}{{\color[HTML]{5B9BD5} \textbf{0.276}}} & 0.613$^*$ & \multicolumn{1}{c|}{0.359$^*$} & 0.577 & \multicolumn{1}{c|}{0.321} & 0.466 & \multicolumn{1}{c|}{0.296} & 0.530 & \multicolumn{1}{c|}{0.293} & 0.617 & \multicolumn{1}{c|}{0.336} & 0.598 & \multicolumn{1}{c|}{0.370} & 0.789 & \multicolumn{1}{c|}{0.505} & 0.604 & \multicolumn{1}{c|}{0.373} & 0.613 & \multicolumn{1}{c|}{0.340} & 0.616 & \multicolumn{1}{c|}{0.382} & 0.649 & 0.438 \\

\multicolumn{1}{c|}{} & \textbf{336} & {\color[HTML]{5B9BD5} \textbf{0.440}} & \multicolumn{1}{c|}{{\color[HTML]{FF0000} \textbf{0.272}}} 
& 0.472 & 0.313 & 0.498 & 0.296
& 0.508 & \multicolumn{1}{c|}{0.324} & 0.486 & \multicolumn{1}{c|}{0.325} & {\color[HTML]{FF0000} \textbf{0.433}} & \multicolumn{1}{c|}{{\color[HTML]{5B9BD5} \textbf{0.283}}} & 0.642$^*$ & \multicolumn{1}{c|}{0.376$^*$} & 0.588 & \multicolumn{1}{c|}{0.324} & 0.482 & \multicolumn{1}{c|}{0.304} & 0.558 & \multicolumn{1}{c|}{0.305} & 0.629 & \multicolumn{1}{c|}{0.336} & 0.605 & \multicolumn{1}{c|}{0.373} & 0.797 & \multicolumn{1}{c|}{0.508} & 0.621 & \multicolumn{1}{c|}{0.383} & 0.618 & \multicolumn{1}{c|}{0.328} & 0.622 & \multicolumn{1}{c|}{0.337} & 0.653 & 0.472 \\

\multicolumn{1}{c|}{} & \textbf{720} & {\color[HTML]{FF0000} \textbf{0.459}} & \multicolumn{1}{c|}{{\color[HTML]{5B9BD5} \textbf{0.308}}} 
& 0.508 & 0.332 & 0.506 & 0.313
& 0.546 & \multicolumn{1}{c|}{0.341} & 0.524 & \multicolumn{1}{c|}{0.348} & {\color[HTML]{5B9BD5} \textbf{0.467}} & \multicolumn{1}{c|}{{\color[HTML]{FF0000} \textbf{0.302}}} & 0.702$^*$ & \multicolumn{1}{c|}{0.401$^*$} & 0.597 & \multicolumn{1}{c|}{0.337} & 0.514 & \multicolumn{1}{c|}{0.322} & 0.589 & \multicolumn{1}{c|}{0.328} & 0.640 & \multicolumn{1}{c|}{0.350} & 0.645 & \multicolumn{1}{c|}{0.394} & 0.841 & \multicolumn{1}{c|}{0.523} & 0.626 & \multicolumn{1}{c|}{0.382} & 0.653 & \multicolumn{1}{c|}{0.355} & 0.660 & \multicolumn{1}{c|}{0.408} & 0.639 & 0.437 \\ \cmidrule(l){2-36} 

\multicolumn{1}{c|}{\multirow{-5}{*}{\textbf{Traffic}}} & \textbf{Avg} & {\color[HTML]{FF0000} \textbf{0.423}} & \multicolumn{1}{c|}{{\color[HTML]{FF0000} \textbf{0.276}}} 
& 0.471 & 0.315 & 0.484 & 0.297
& 0.514 & \multicolumn{1}{c|}{0.328} & 0.483 & \multicolumn{1}{c|}{0.325} & {\color[HTML]{5B9BD5} \textbf{0.428}} & \multicolumn{1}{c|}{{\color[HTML]{5B9BD5} \textbf{0.282}}} & 0.641$^*$ & \multicolumn{1}{c|}{0.370$^*$} & 0.583 & \multicolumn{1}{c|}{0.323} & 0.481 & \multicolumn{1}{c|}{0.304} & 0.550 & \multicolumn{1}{c|}{0.304} & 0.620 & \multicolumn{1}{c|}{0.336} & 0.625 & \multicolumn{1}{c|}{0.383} & 0.804 & \multicolumn{1}{c|}{0.509} & 0.610 & \multicolumn{1}{c|}{0.376} & 0.624 & \multicolumn{1}{c|}{0.340} & 0.628 & \multicolumn{1}{c|}{0.379} & 0.650 & 0.446 \\ \midrule

\multicolumn{1}{c|}{} & \textbf{96} & {\color[HTML]{5B9BD5} \textbf{0.162}} & \multicolumn{1}{c|}{0.214} 
& {\color[HTML]{5B9BD5} \textbf{0.162}} & {\color[HTML]{FF0000} \textbf{0.207}} & 0.163 & 0.209
& 0.167 & \multicolumn{1}{c|}{0.212} & 0.171 & \multicolumn{1}{c|}{0.220} & 0.174 & \multicolumn{1}{c|}{0.214} & 0.163 & \multicolumn{1}{c|}{0.212} & 0.159 & \multicolumn{1}{c|}{0.218} & 0.177 & \multicolumn{1}{c|}{0.218} & {\color[HTML]{FF0000} \textbf{0.158}} & \multicolumn{1}{c|}{0.230} & 0.172 & \multicolumn{1}{c|}{0.220} & 0.196 & \multicolumn{1}{c|}{0.255} & 0.221 & \multicolumn{1}{c|}{0.306} & 0.217 & \multicolumn{1}{c|}{0.296} & 0.173 & \multicolumn{1}{c|}{0.223} & 0.266 & \multicolumn{1}{c|}{0.336} & 0.230 & 0.329 \\

\multicolumn{1}{c|}{} & \textbf{192} & 0.214 & \multicolumn{1}{c|}{0.258} 
& 0.209 & {\color[HTML]{5B9BD5} \textbf{0.252}} & {\color[HTML]{5B9BD5} \textbf{0.208}} & {\color[HTML]{FF0000} \textbf{0.250}}
& 0.213 & \multicolumn{1}{c|}{0.253} & 0.230 & \multicolumn{1}{c|}{0.266} & 0.221 & \multicolumn{1}{c|}{0.254} & 0.212 & \multicolumn{1}{c|}{0.254} & 0.211 & \multicolumn{1}{c|}{0.266} & 0.225 & \multicolumn{1}{c|}{0.259} & {\color[HTML]{FF0000} \textbf{0.206}} & \multicolumn{1}{c|}{0.277} & 0.219 & \multicolumn{1}{c|}{0.261} & 0.237 & \multicolumn{1}{c|}{0.296} & 0.261 & \multicolumn{1}{c|}{0.340} & 0.276 & \multicolumn{1}{c|}{0.336} & 0.245 & \multicolumn{1}{c|}{0.285} & 0.307 & \multicolumn{1}{c|}{0.367} & 0.263 & 0.322 \\

\multicolumn{1}{c|}{} & \textbf{336} & {\color[HTML]{FF0000} \textbf{0.251}} & \multicolumn{1}{c|}{{\color[HTML]{5B9BD5} \textbf{0.290}}}
& {\color[HTML]{5B9BD5} \textbf{0.263}} & 0.292 & {\color[HTML]{FF0000} \textbf{0.251}} & {\color[HTML]{FF0000} \textbf{0.287}}
& 0.268 & \multicolumn{1}{c|}{0.292} & 0.280 & \multicolumn{1}{c|}{0.299} & 0.278 & \multicolumn{1}{c|}{0.296} & 0.272 & \multicolumn{1}{c|}{0.299} & 0.267 & \multicolumn{1}{c|}{0.310} & 0.278 & \multicolumn{1}{c|}{0.297} & 0.272 & \multicolumn{1}{c|}{0.335} & 0.280 & \multicolumn{1}{c|}{0.306} & 0.283 & \multicolumn{1}{c|}{0.335} & 0.309 & \multicolumn{1}{c|}{0.378} & 0.339 & \multicolumn{1}{c|}{0.380} & 0.321 & \multicolumn{1}{c|}{0.338} & 0.359 & \multicolumn{1}{c|}{0.395} & 0.354 & 0.396 \\

\multicolumn{1}{c|}{} & \textbf{720} & {\color[HTML]{FF0000} \textbf{0.333}} & \multicolumn{1}{c|}{{\color[HTML]{FF0000} \textbf{0.338}}}
& 0.344 & 0.344 & {\color[HTML]{5B9BD5} \textbf{0.339}} & {\color[HTML]{5B9BD5} \textbf{0.341}} &
0.345 & \multicolumn{1}{c|}{0.341} & 0.352 & \multicolumn{1}{c|}{0.344} & 0.358 & \multicolumn{1}{c|}{0.347} & 0.350 & \multicolumn{1}{c|}{0.348} & 0.352 & \multicolumn{1}{c|}{0.362} & 0.354 & \multicolumn{1}{c|}{0.348} & 0.398 & \multicolumn{1}{c|}{0.418} & 0.365 & \multicolumn{1}{c|}{0.359} & 0.345 & \multicolumn{1}{c|}{0.381} & 0.377 & \multicolumn{1}{c|}{0.427} & 0.403 & \multicolumn{1}{c|}{0.428} & 0.414 & \multicolumn{1}{c|}{0.410} & 0.419 & \multicolumn{1}{c|}{0.428} & 0.409 & 0.371 \\ \cmidrule(l){2-36} 

\multicolumn{1}{c|}{\multirow{-5}{*}{\textbf{Weather}}} & \textbf{Avg} & {\color[HTML]{FF0000} \textbf{0.240}} & 0.275
& 0.244 & {\color[HTML]{5B9BD5} \textbf{0.274}} & {\color[HTML]{FF0000} \textbf{0.240}} & \color[HTML]{FF0000} \textbf{0.271}
& 0.248 & \multicolumn{1}{c|}{{0.277}} & 0.258 & \multicolumn{1}{c|}{0.282} & 0.258 & \multicolumn{1}{c|}{0.278} & 0.249 & \multicolumn{1}{c|}{0.278} & 0.247 & \multicolumn{1}{c|}{0.289} & 0.259 & \multicolumn{1}{c|}{0.281} & 0.259 & \multicolumn{1}{c|}{0.315} & 0.259 & \multicolumn{1}{c|}{0.287} & 0.265 & \multicolumn{1}{c|}{0.317} & 0.292 & \multicolumn{1}{c|}{0.363} & 0.309 & \multicolumn{1}{c|}{0.360} & 0.288 & \multicolumn{1}{c|}{0.314} & 0.338 & \multicolumn{1}{c|}{0.382} & 0.314 & 0.355 \\ \midrule

\multicolumn{2}{c|}{\textbf{Count}} & {\color[HTML]{FF0000} \textbf{26}} & \multicolumn{1}{c|}{{\color[HTML]{FF0000}\textbf{26}}} 
& \textbf{1} & \textbf{6} & \textbf{7} & \textbf{6}
& \textbf{2} & \multicolumn{1}{c|}{ \textbf{1}} & \textbf{0} & \multicolumn{1}{c|}{\textbf{0}} & \textbf{1} & \multicolumn{1}{c|}{2} & \textbf{0} & \multicolumn{1}{c|}{\textbf{0}} & \textbf{1} & \multicolumn{1}{c|}{\textbf{0}} & \textbf{1} & \multicolumn{1}{c|}{\textbf{1}} & \textbf{2} & \multicolumn{1}{c|}{\textbf{0}} & \textbf{1} & \multicolumn{1}{c|}{\textbf{0}} & \textbf{0} & \multicolumn{1}{c|}{\textbf{0}} & \textbf{0} & \multicolumn{1}{c|}{\textbf{0}} & \textbf{1} & \multicolumn{1}{c|}{\textbf{0}} & \textbf{0} & \multicolumn{1}{c|}{\textbf{0}} & \textbf{0} & \multicolumn{1}{c|}{\textbf{0}} & \textbf{0} & \textbf{0} \\ \bottomrule
\end{tabular}}
\end{table*}

\noindent \textbf{Baselines.}
We choose \textbf{16 competitive baselines}.
(1) Variable-independent methods:
Autoformer \cite{wu2021autoformer} decomposes time series into seasonal and trend components and leverages auto-correlation for series-level aggregation. 
FEDformer \cite{zhou2022fedformer} enhances decomposition with frequency-based Transformers and kernel moving averages. 
PatchTST \cite{nie2022time} adopts channel-independent patching, treating subseries as tokens for Transformer-based semantic extraction.
Stationary \cite{liu2022non} addresses predictability–capacity tradeoffs via stationarization and de-stationary attention. 
DLinear \cite{zeng2023transformers} linearly models trend and seasonal components. 
SCINet \cite{liu2022scinet} applies a downsample–convolve–interact design for multi-resolution temporal features. 
TimesNet \cite{wu2022timesnet} captures multi-periodicity by mapping 1D series to 2D tensors. 
TimeMixer \cite{wang2024timemixer} employs an MLP-based architecture with decomposable past- and multi-predictor future-mixing blocks.
(2) Variable-dependent methods:
MTGNN \cite{mtgnn} combines mix-hop and graph learning to capture spatial–temporal dependencies. Crossformer \cite{zhang2023Crossformer} leverages cross-dimension attention for inter-series dependencies. CrossGNN \cite{huang2023crossgnn} builds multi-edge graphs and multi-patch encodings for spatial and multi-scale temporal dependencies. MSGNet \cite{cai2024msgnet} integrates multi-head attention with adaptive graph convolutions for intra- and inter-variate correlations. iTransformer \cite{liuitransformer} applies attention and feed-forward networks on inverted dimensions for inter-variate dependencies. FilterTS \cite{wang2025filterts} uses static and dynamic filters to enhance frequency component extraction in MTS.
% MTGNN \cite{mtgnn} integrates a mix-hop mechanism with a graph learning module to capture both spatial and temporal dependencies within the time series.
% Crossformer \cite{zhang2023Crossformer} utilizes cross-dimension attention to capture inter-series dependencies in multivariate time series forecasting.
% CrossGNN \cite{huang2023crossgnn} constructs multi-edge graphs to capture spatial dependencies and uses a multi-patch encoding method to capture multi-scale temporal dependencies.
% MSGNet \cite{cai2024msgnet} combines multi-head attention with adaptive graph convolution modules to capture intra-variate and inter-variate correlations on different time scales.
% iTransformer \cite{liuitransformer} applies attention and feed-forward networks to inverted dimensions to capture inter-variate dependencies.
% \wu{FilterTS \cite{wang2025filterts} leverages static and dynamic filters to enhance MTS frequency component extraction.}
(3) Single-scale lead-lag methods:
LIFT \cite{zhao2024rethinking} models lead-lag dependencies via cross-correlation and integrates a frequency-based mixer to combine leading information. As a plug-and-play post-processing module, it is not directly comparable. Thus, we adopt LightMTS, its official variant that integrates LIFT on a linear backbone, for performance evaluation. 
Vcformer \cite{yang2024vcformer} calculates cross-correlation for different time lags between queries and keys, integrating them to capture lead-lag dependencies.

\noindent \textbf{Implement Details}
All experiments are implemented in PyTorch and conducted on a single NVIDIA A100 80GB GPU. We use the Adam optimizer with initial learning rates of $\{10^{-2}, 10^{-3}, 5 \times 10^{-4},$ $10^{-4} \}$. The batch size is set to 32, and the number of training epochs is set to 10 with each stopping tolerance of 3.
To turn hyperparameters of MillGNN, we use an open source AutoML toolkit NNI \cite{nni2021} to automatically search for optimal hyperparameters while significantly reducing computational costs. Specifically, we employ its built-in Bayesian optimization algorithm as the tuner and set the maximum number of trials to 20. In addition, we use the built-in Curve Fitting assessor to terminate trials early if the learning curve is likely to converge to a suboptimal result. 
Considering that variate numbers affect multi-scale inter-variate structures, we design two settings with 200 variates as the threshold. \textbf{Setting 1} (ETTh1/2, ETTm1/2, Exchange, Weather; $\leq$21 variates) searches scales in \{1,2\} with groups at scale 1 from \{1,2,3\}, while scale 2 has no grouping. \textbf{Setting 2} (Electricity, Traffic, PEMS04/08, China-AQI; $\geq$170 variates) searches scales in \{1,2,3\}, groups at scale 1 from \{5,10,15,20,25\}, and groups at scale 2 from \{1,2,3\}. For all datasets, patch length $\in$ \{2,4,8,12,24\} and selected time lags $\in$ \{4,6,8,10,15,20\}.
% Given that the number of variates influences the construction of multi-scale inter-variate structures, and datasets differ in the number of variates, we design two experimental settings, using 200 variates as the dividing criterion.
% The hyperparameter search space is shown in Table \ref{hyper}.

\subsection{Main Results}
Table \ref{tab:long-term} and Table \ref{tab:short-term} summarize the long-term and short-term forecasting results, respectively. Red and Blue denote the best and second best performance, respectively. The results with $*$ are rerun by us to meet our settings using their official codes. For those without $*$ are cited from iTransformer \cite{liuitransformer}, TimeMixer \cite{wang2024timemixer}, or the respective original papers.

\begin{table}[]
\centering
\caption{Short-term forecasting results with the input length $L=96$ and predicted lengths $H=12$ for PEMS04/08 and $H=24$ for China AQI.}
\label{tab:short-term}
\resizebox{0.98 \linewidth}{!}{
\begin{tabular}{c|ccc|ccc|ccc}
\toprule
Dataset & \multicolumn{3}{c|}{PEMS04} & \multicolumn{3}{c|}{PEMS08} & \multicolumn{3}{c}{China AQI$^*$} \\
\midrule
Metric & MAE & MAPE & RMSE & MAE & MAPE & RMSE & MAE & MAPE & RMSE \\
\midrule
DLinear	& 24.62 & 16.12	&39.51&	20.26	&12.09	&32.38 & 24.18 & 36.33 & 36.89 \\
TimesNet &	21.63	& 13.15 &	34.90 &	19.01 &	11.83 &	30.65 & 23.40 & 35.85 & 36.12 \\
Crossformer & 20.38 & 12.84 & 32.41 & 17.56 & 10.92 & 27.21 & 21.22 & 34.97 & 34.43 \\
PatchTST & 24.86 & 16.65 & 40.46 & 20.35 & 13.15 & 31.04 & 23.17 & 34.56 & 35.33 \\
CrossGNN$^*$ &  19.33 & 12.46 & 31.31 & 15.42 & 9.84 & 23.91 & {\ul 19.81} & {\ul 32.34} & {\ul 32.60} \\
MSGNet$^*$ & 22.03 & 12.92 & 34.24 & 16.32 & 10.31 & 25.49 & 22.09 & 34.58 & 35.76 \\
iTansformer$^*$ & 20.16 & {\ul12.40} & 31.64 & 16.01 & 10.05 & 25.13 & 20.32 & 33.45 & 34.37 \\
LIFT $^*$ & 19.41 & 12.53 & 32.35 & {\ul 15.22} & 10.18 & 23.86 & 20.14 & {\ul32.34}  & 32.66  \\
VCformer$^*$  & 19.87 & 12.46 & 31.56 & 15.30 & 10.09 & {\ul23.84} & 20.36 & 33.35 & 33.97  \\
\wu{TimeMixer} & {\ul19.21} & 12.53 & {\ul 30.92} & {\ul15.22} & {\ul 9.67} & 24.26 & 21.60 & 34.30 & 35.19 \\
\wu{FilterTS}$^*$ & 20.49 & 12.76 & 31.16 & 15.88 & 10.28 & 24.64 & 21.35 & 34.73 & 34.27 \\
\midrule
MillGNN & \textbf{18.93} & \textbf{12.24} & \textbf{30.72} & \textbf{14.86} & \textbf{9.54} & \textbf{23.46} & \textbf{19.05} & \textbf{31.51} & \textbf{31.55} \\
\bottomrule
\end{tabular}}
\end{table}
% the following tendencies can be discerned

From these results, the following tendencies can be discerned:

(1) \textbf{MillGNN vs. Variate-Independent/Dependent Methods}: 
MillGNN outperforms competitive variate-independent/dependent methods, as MillGNN can effectively learn lead-lag dependencies in addition to intra-variate and time-aligned inter-variate dependencies. This ability enables MillGNN to comprehensively consider synchronous and asynchronous effects across intra- and inter-variates.

(2) \textbf{MillGNN vs. Single-Scale Lead-Lag Methods}: 
MillGNN surpasses methods that model single-scale lead-lag dependencies, as MillGNN can learn lead-lag dependencies at multiple grouping scales.
This ability enables MillGNN to capture delays not only between variates but also between variate groups.

(3) \textbf{Impacts of Correlations between Variates}:
For datasets that exhibit strong spatial correlations between variates, MillGNN delivers pronounced gains. Specifically, for the Traffic dataset, MillGNN outperforms the best baselines with an average MSE of 2.7\%. For the PEMS04, PEMS08, and China-AQI datasets, MillGNN outperforms the best baselines with an average MAE of 2.1\%, 2.4\%, and 3.8\%, respectively. This may be because the propagation of traffic flow and air pollution is influenced by distance, resulting in hierarchical patterns of propagation. Areas closer to the source tend to experience more significant impacts. This phenomenon aligns with the hierarchical message-passing design of MillGNN.

\begin{figure}
    \centering
    \includegraphics[width= \linewidth]{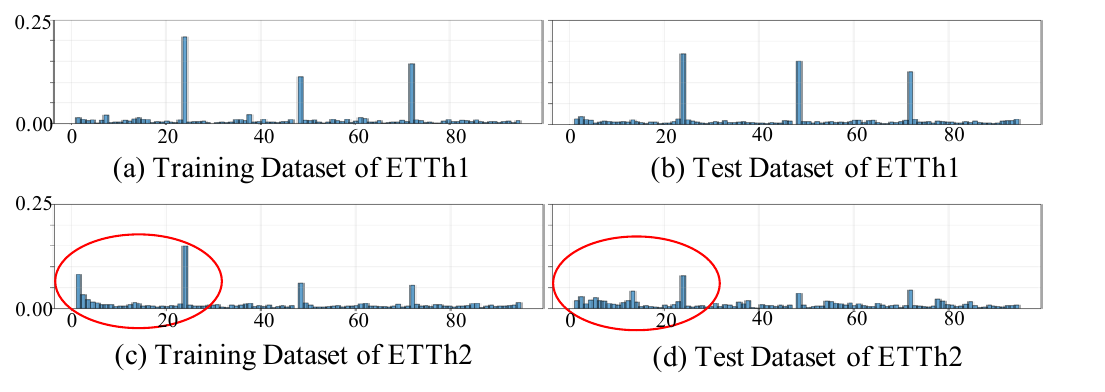}
    \caption{Lag distributions of the training and test datasets on the ETTh1 and ETTh2 datasets}
    \label{fig:distribution}
\end{figure}

(3) \textbf{Impacts of Lag Distribution Drifts}:
On the ETTh1 and ETTh2 datasets, which both record hourly Electricity Transformer Temperature data from one oil temperature sensor and six power load sensors, MillGNN performs best on ETTh1 in 7 out of 8 cases but slightly underperforms on ETTh2. To investigate the reason, Fig. \ref{fig:distribution} illustrates the lag distributions in the ETTh1 and ETTh2 datasets, derived via cross-correlation. The lag distributions in the ETTh1 dataset are similar between training and test sets, whereas those in ETTh2 show notable differences, particularly in the red-circled region. These results suggest that MillGNN performs better on datasets with stable lag distributions between training and test sets, providing empirical guidance for its application.

\subsection{Ablation Study}
\begin{table}[]
\centering
\caption{Results of removing key components.}
\label{tab:abl}
\resizebox{0.9 \linewidth}{!}{
\begin{tabular}{c|c|cccc|c}
\toprule
Dataset & Metric & w/o ms & w/o init & w/o weight & w/o hmp & MilGNN \\
\midrule
\wu{ETTm1} & MSE & 0.379 & 0.388 & 0.394 & 0.385 & \textbf{0.373} \\
% Electricity & MAE & 0.271 & 0.278 & 0.281 & 0.273 & \textbf{0.263}\\
Electricity & MSE & 0.176 & 0.182 & 0.183 & 0.179 & \textbf{0.170} \\
% \midrule
% \wu{PEMS08} & MAE & 15.22 & 15.28 & 15.33 & 15.25 & \textbf{15.17}\\    
China-AQI & MAE & 19.61 & 19.88 & 20.02 & 19.53 & \textbf{19.05} \\
% (209 variables)& RMSE & 32.17 & 32.24 & 32.44 & 32.22& \textbf{31.55} \\
\bottomrule
\end{tabular}}
\end{table}

\noindent \textbf{Impacts of Key Components.}
We remove key components of MillGNN to investigate their effectiveness:
(1) w/o multi-scale construction (w/o ms): Removing multiple grouping scales, which only model lead-lag dependencies between variates.
(2) w/o initial lead-lag graphs $\{\boldsymbol{C}^s\}_{s=0}^{S-1}$ (w/o init): Modeling multi-scale lead-lag dependencies purely on learnable weights $\{\boldsymbol{\Lambda}^s\}_{s=0}^{S-1}$ without cross-correlations.
(3) w/o decay-aware attention machanism $\{\boldsymbol{\Lambda}^s\}_{s=0}^{S-1}$ (w/o weight): Modeling multi-scale lead-lag dependencies purely on $\{\boldsymbol{C}^s\}_{s=0}^{S-1}$ without dynamic adjustment.
(4) w/o HiLL-MP module (w/o hmp): Passing lead-lag messages using Vanilla GNN \cite{kipf2017semi}.

The results are summarized in Table \ref{tab:abl}, where MillGNN performs best, validating the effectiveness of its key components. Specifically, 
(1) Removing lead-lag graphs and dynamic decaying weights leads to performance drops, highlighting the importance of our strategy, i.e., learning evolving dependencies under the statistic guidance. Notably, removing dynamic decaying weights results in a more pronounced performance decline, especially on the China-AQI dataset. This is likely due to the diffusion characteristics of air pollution, which spread over hours between cities or days between areas. These results highlight MillGNN's strength for modeling dynamics and decay.
(2) Omitting the multi-scale construction and hierarchical lead-lag message passing module results in performance degradation. This underscores the critical role of capturing lead-lag dependencies at multiple grouping scales and validates our motivation to learn and utilize these dependencies.

\begin{table}[]
\centering
\caption{Results of different clustering algorithms.}
\label{tab:variation}
\resizebox{0.75\linewidth}{!}{
\begin{tabular}{c|c|ccc}
\toprule
Dataset & Metric & \multicolumn{1}{c}{Hierarchical} & \multicolumn{1}{c}{K-means} & \multicolumn{1}{c}{Spectral(Ours)} \\ \midrule
Electricity & MSE & 0.194 & 0.176 & \textbf{0.170}  \\
PEMS04 & MAE & 21.08 & 19.34 & \textbf{18.93} \\
\bottomrule
\end{tabular}}
\end{table}

\begin{table}[]
\centering
\caption{Results of different inter-scale passing designs on the China-AQI dataset.}
\label{tab:inter_scale}
\resizebox{0.8\linewidth}{!}{
\begin{tabular}{c|ccc}
\toprule
Metrics & Learnable & Bi-direction & Duplication (Ours)\\
\midrule
GPU Memory (GB) & 6.44 & 7.28 & \textbf{5.62}\\
Training Time (min/epoch) & 2.07 & 4.55 & \textbf{1.49}\\
MAE & \textbf{18.94} & 20.13 & 19.05 \\
\bottomrule
\end{tabular}}
\end{table}

\noindent \wu{\textbf{Impacts of Different Clustering Algorithms.}
Since MillGNN can ingest any variate‑to-group scheme, we conduct variations with three popular clustering algorithms, i.e., hierarchical, K-means, and spectral, to investigate the effects. From Table \ref{tab:variation}, we can observe that spectral clustering yields the best accuracy. However, its eigen‑decomposition is empirically time-consuming. On large‑scale datasets, where runtime becomes a bottleneck, K‑means strikes a more favorable accuracy–efficiency trade‑off.}

\noindent \wu{\textbf{Impacts of Inter-Scale Message-Passing Strategies.}
We evaluated three variants, including 
(1) duplicate (cause-to-fine): 
Messages are duplicated across the pre‑computed assignment matrix.
(2) weight: Messages are re‑weighted via a learnable assignment matrix \cite{ying2018hierarchical}.
(3) bi‑direction: The weighted scheme in (2) is extended to send messages in both directions.
From Table \ref{tab:inter_scale}, we can observe that the bi-direction variation performs worst with the highest computation cost. This may be because the complicated interaction over-mixes representations and washes out salient cues. The weight variation improves accuracy slightly but inflates training time by around 1.4 times compared with the duplicate variation. 
Our duplication design balances effectiveness and efficiency, which is more practical.}

\begin{figure}
    \centering
    \includegraphics[width=\linewidth]{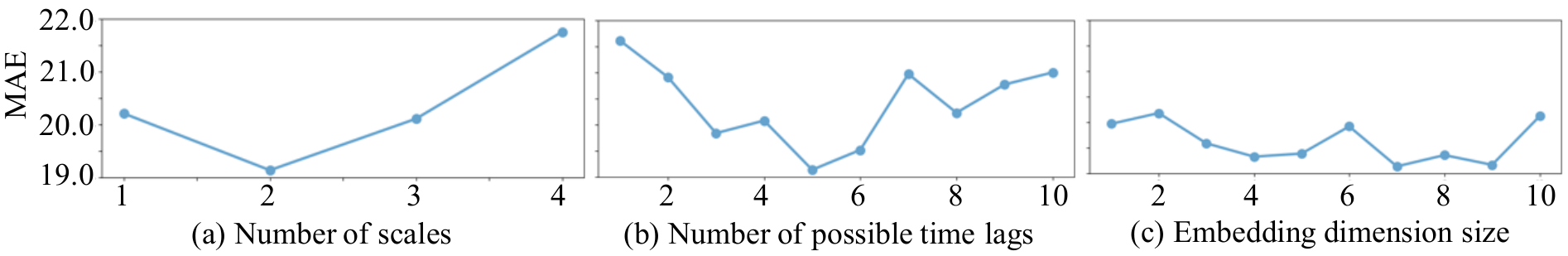}
    \caption{Results of the hyperparameter study on the China-AQI dataset. }
    \label{fig:hyper}
\end{figure}

\subsection{Hyperparameter Study}
We conduct a hyperparameter study on the China-AQI dataset to explore the effects of key hyperparameters. These hyperparameters include the number of grouping scales $S$, the number of possible time lags between pairwise groups $K$, and the embedding dimension $d_{\text{e}}$ for learning lead-lag weights. The results are shown in Fig. \ref{fig:hyper}. We observe that:
(1) $S$ (optimal $S=2$) and $K$ (optimal $K=5$) play critical roles in MillGNN’s performance, as they determine the total number of lead-lag dependencies at multiple grouping scales. With increasing $S$ and $K$, the number of multi-scale lead-lag dependencies grows, leading to a performance improvement followed by a decline. The reason may be that insufficient multi-scale lead-lag dependencies fail to comprehensively address lead-lag effects, while an excessive number of dependencies may result in overfitting.
(2) $d_{\text{e}}$ (optimal $d_{\text{e}}=7$) affects the learning of decays. Too many or too few parameters can lead to overfitting or underfitting when modeling the decays.

\subsection{Efficiency Study}

\begin{figure}
    \centering
    \includegraphics[width=0.8 \linewidth]{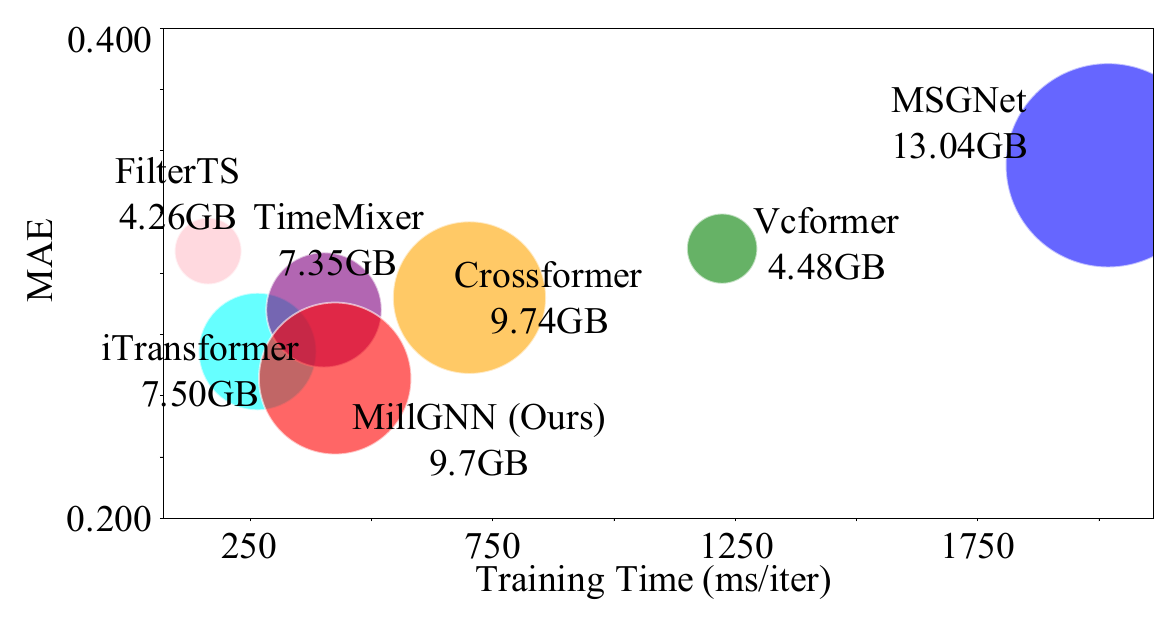}
    \caption{Results of the efficiency study. Efficiency comparison under the input-96-output-96 setting on the Traffic dataset (862 variables).}
    \label{fig:efficiency}
\end{figure}

We evaluate the efficiency by comparing their training time, GPU memory, and accuracy.
We compare MillGNN with competitive baselines, including Crossformer (2023), MSGNet (2024), iTransformer (2024), VCFormer (2024), TimeMixer (2024), and FilterTS (2025).
As illustrated in Fig. \ref{fig:efficiency}, MillGNN achieves the best accuracy while maintaining affordable efficiency. Its accuracy stems from its ability to effectively handle delays between variates and between variate groups. 
For training time and GPU usage, due to modeling multi-scale lead-lag dependencies, MillGNN shows no clear advantage over variate-independent and time-aligned methods.
However, benefiting from hierarchical message passing, which optimizes the passing of multi-scale features via integrating structure and duplication, MillGNN outperforms time-unaligned methods (e.g., Crossformer) and multi-scale methods (e.g., MSGNet) with a gap.

\subsection{Case Study}
\begin{figure}
    \centering
    \includegraphics[width= \linewidth]{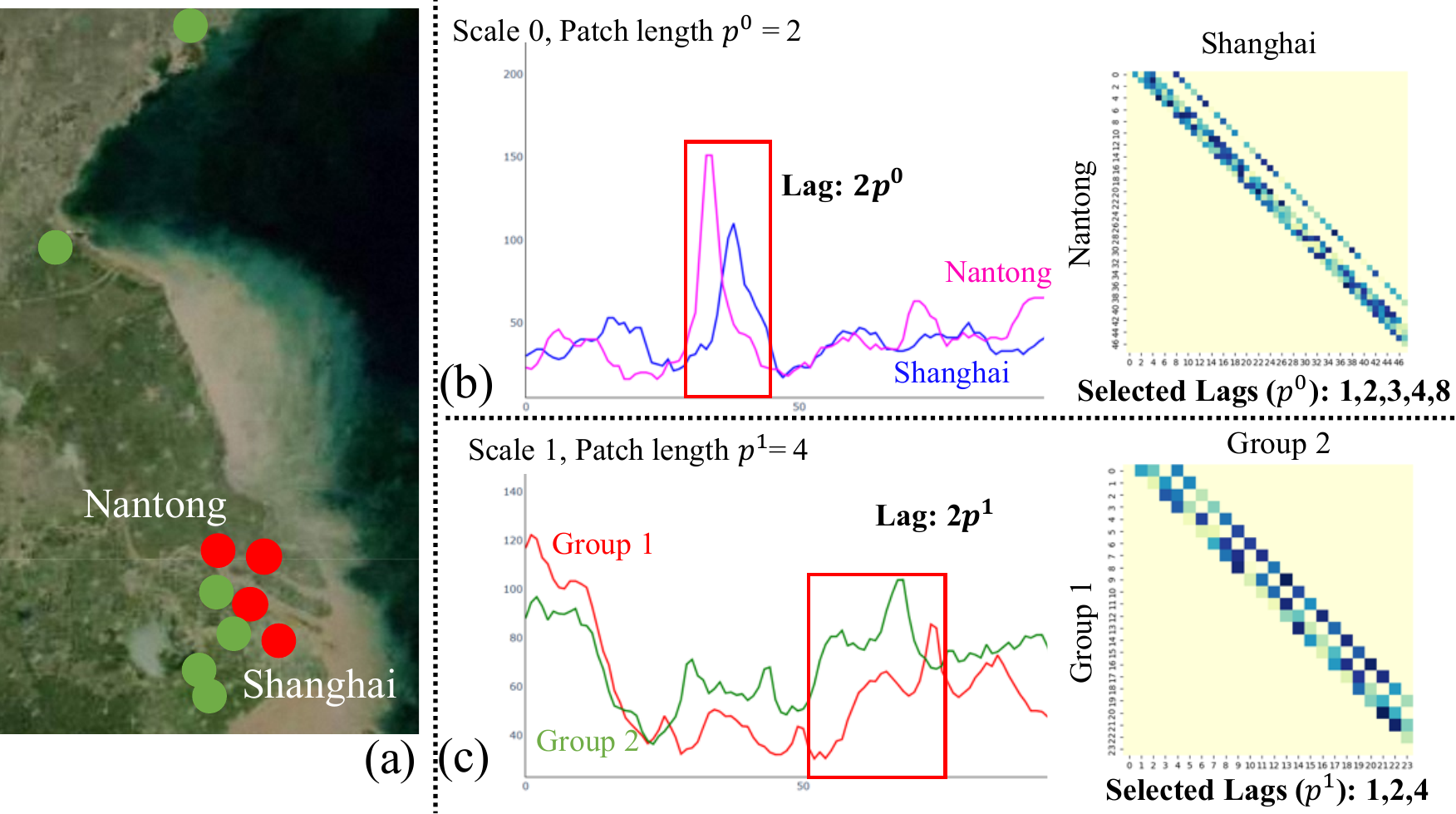}
    \caption{Example of multi-scale lead-lag dependencies on the China AQI dataset.}
    \label{fig:llgraph}
\end{figure}

\textbf{Multi-Scale Lead-Lag Dependencies}
To demonstrate MillGNN's effectiveness in modeling multi-scale lead-lag dependencies, we analyze two variates and two variate groups.
Fig. \ref{fig:llgraph}(a) shows the learned assignment of variates of two groups, indicated in red and green.
Fig. \ref{fig:llgraph}(b) highlights that Nantong and Shanghai, both in the red group, exhibit similar AQI patterns with observable shifts. The learned lead-lag graph identifies Nantong influencing Shanghai with time lags, closely aligning with actual AQI dynamics.
Fig. \ref{fig:llgraph}(c) reveals that the green group exerts lead-lag effects on the red group, with the learned graph capturing time lags, reflecting this cross-group phenomenon.

\begin{figure}
    \centering
    \includegraphics[width= \linewidth]{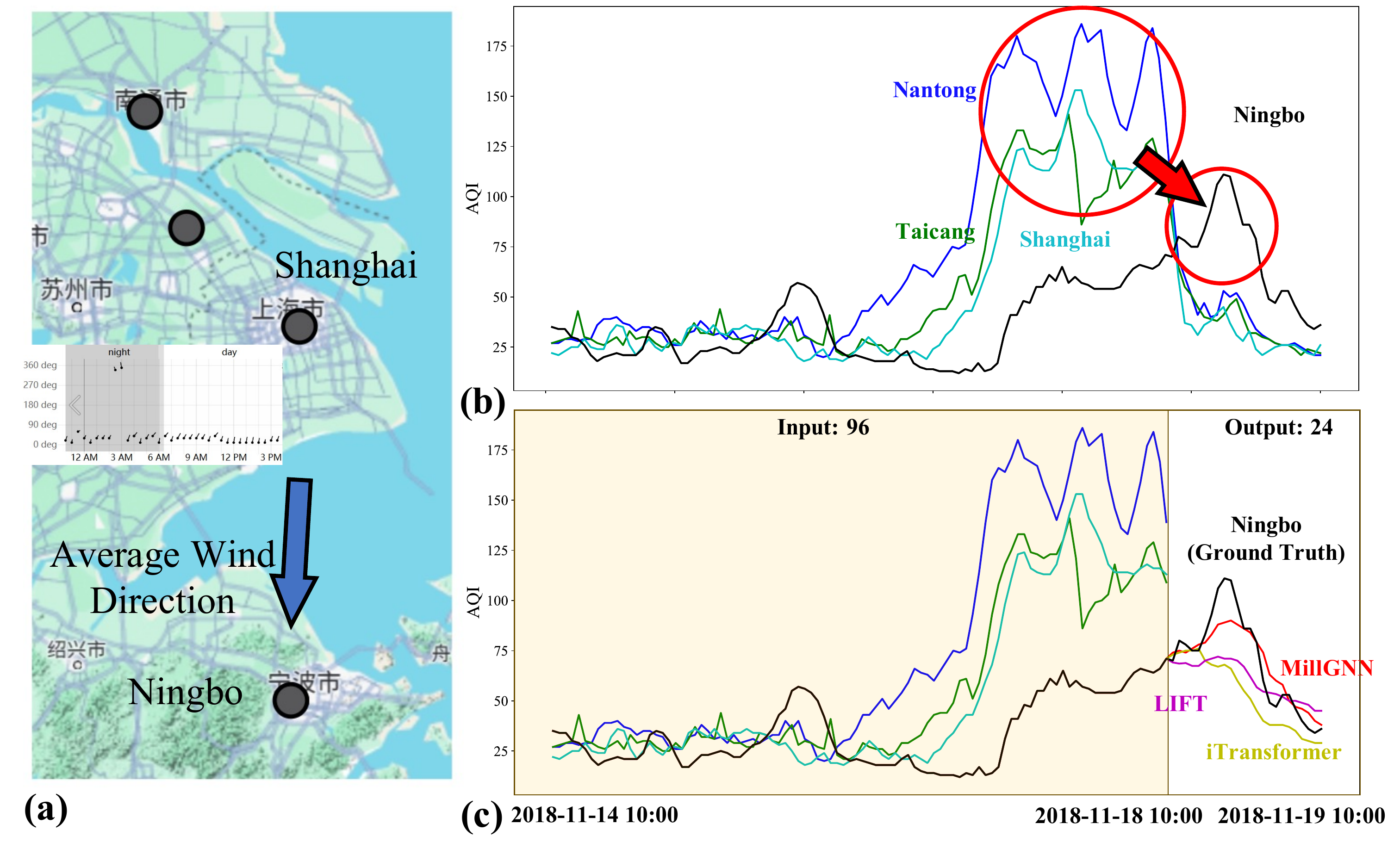}
    \caption{Example of predicting AQI in Ningbo.}
    \label{fig:prediction}
\end{figure}

\noindent \textbf{Real-world AQI Prediction}
We select a case about the predicted results of Ningbo in the test dataset. As shown in Fig. \ref{fig:prediction}, the average wind direction in Shanghai during Nov.17 and Nov. 18 2018 was north and northeast (shown in a), according to the historical data from WeatherSpark \footnote{https://weatherspark.com/}, resulting in air pollution peaks with lags between Shanghai and Ningbo (shown in b). MillGNN can effectively detect these lags purely on AQI data without any wind information, while other SOTA methods cannot (shown in c).

\section{Conclusion and Future Work}
In this work, we propose MillGNN to learn multiple grouping scale lead-lag dependencies for MTS forecasting. We first group variables based on temporal similarity to construct multiple grouping scales. We then introduce a scale-specific module to learn evolving lead-lag dependencies for each scale and a hierarchical message passing module to simultaneously propagate intra- and inter-scale lead-lag effects. Experimental results on 11 real-world MTS datasets demonstrate the SOTA performance of MillGNN. 
In future work, we will discover more complex lead-lag dependencies, e.g., high-order and sensitive to data drifts, to better adapt to diverse real-world scenarios. In addition, we will explore efficient and differentiable grouping strategies, e.g., grouping based on trend-seasonal characteristics, for non-uniform inter-scale weighting mechanisms.

\bibliographystyle{ACM-Reference-Format}
\bibliography{reference}

\appendix

\end{document}